\documentclass[final,3p,times,twocolumn]{elsarticle}

\usepackage{amsmath,amsfonts,bm,amsthm,amssymb, bbding}
\usepackage{algorithmic}
\usepackage{algorithm}
\usepackage{array}
\usepackage{textcomp}
\usepackage{stfloats}
\usepackage{url}
\usepackage{verbatim}
\usepackage{amsmath,graphicx}
\usepackage{multirow}
\usepackage{makecell}
\usepackage{color}
\usepackage{arydshln}
\usepackage{cite}
\usepackage{algorithm, algorithmic}
\usepackage{amssymb}
\usepackage{booktabs}
\usepackage{pifont}% http://ctan.org/pkg/pifont
\newcommand{\cmark}{\ding{51}}%
\newcommand{\xmark}{\ding{55}}%
\usepackage{diagbox}
\usepackage{multirow}
\journal{Neurocomputing}

\begin{document}

\begin{frontmatter}



\title{TransRef: Multi-Scale Reference Embedding Transformer for Reference-Guided Image Inpainting}


\author[inst1]{Taorong~Liu}
 % are with the School of Computer Science, Wuhan University, Wuhan 430072, China 
\affiliation[inst1]{
organization={School of Computer Science},
addressline={Wuhan University}, 
city={Wuhan},
postcode={430072}, 
state={Hubei},
country={China}
}

\author[inst2]{Liang~Liao}

\affiliation[inst2]{
organization={School of Computer Science and Engineering},%Department and Organization
            addressline={Nanyang Technological University}, 
            city={Singapore},
            postcode={639798}, 
            % state={State Two},
            country={Singapore}}

\affiliation[inst3]{
organization={ Department of Electrical Engineering and the Institute of Communications Engineering},%Department and Organization
            addressline={National Tsing Hua University}, 
            city={Hsinchu},
            postcode={30013}, 
            state={Taiwan},
            country={China}}

\affiliation[inst4]{
organization={Digital Content and Media Sciences Research Division},%Department and Organization
            addressline={National Institute of Informatics}, 
            city={Tokyo},
            postcode={101-8430}, 
            % state={Taiwan},
            country={Japan}}

\author[inst1]{Delin~Chen}
\author[inst1]{Jing~Xiao\corref{cor1}}
\cortext[cor1]{Corresponding Author}

\author[inst1]{Zheng~Wang}
\author[inst3]{Chia-Wen~Lin}
\author[inst4]{Shin'ichi~Satoh}

\begin{abstract}
Image inpainting for completing complicated semantic environments and diverse hole patterns of corrupted images is challenging even for state-of-the-art learning-based inpainting methods trained on large-scale data. A reference image capturing the same scene of a corrupted image offers informative guidance for completing the corrupted image as it shares similar texture and structure priors to that of the holes of the corrupted image. In this work, we propose a \textbf{Trans}former-based encoder-decoder network for \textbf{Ref}erence-guided image inpainting, named \textbf{TransRef}. Specifically, the guidance is conducted progressively through a reference embedding procedure, in which the referencing features are subsequently aligned and fused with the features of the corrupted image. For precise utilization of the reference features for guidance, a reference-patch alignment (Ref-PA) module is proposed to align the patch features of the reference and corrupted images and harmonize their style differences, while a reference-patch transformer (Ref-PT) module is proposed to refine the embedded reference feature. Moreover, to facilitate the research of reference-guided image restoration tasks, we construct a publicly accessible benchmark dataset containing 50K pairs of input and reference images. Both quantitative and qualitative evaluations demonstrate the efficacy of the reference information and the proposed method over the state-of-the-art methods in completing complex holes. Code and dataset can be accessed at: \url{https://github.com/Cameltr/TransRef}.
\end{abstract}

\begin{keyword}
Image inpainting \sep reference embedding \sep feature alignment.
% %% MSC codes here, in the form: \MSC code \sep code
% %% or \MSC[2008] code \sep code (2000 is the default)
\end{keyword}

\end{frontmatter}


\section{Introduction}
\label{sec:introduction}
Image inpainting refers to the process of filling the missing areas of an image with plausible content. This task has been an active topic in image processing for decades, since it has found a broad range of applications, such as object removal, image editing, error concealment in video transmission, and hole filling in 3D reconstruction. The key to producing high-quality image inpainting results lies in finding sufficient priors such as smoothness~\citep{bertalmio2000image,1401905}, self-similarity~\citep{18,PatchMatch}, and sparse~\citep{TIPxianming,sparsetip}, to serve as the inpainting guide for hole filling.

\begin{figure}[t]
	\centering
 \includegraphics[width=0.8 \columnwidth]{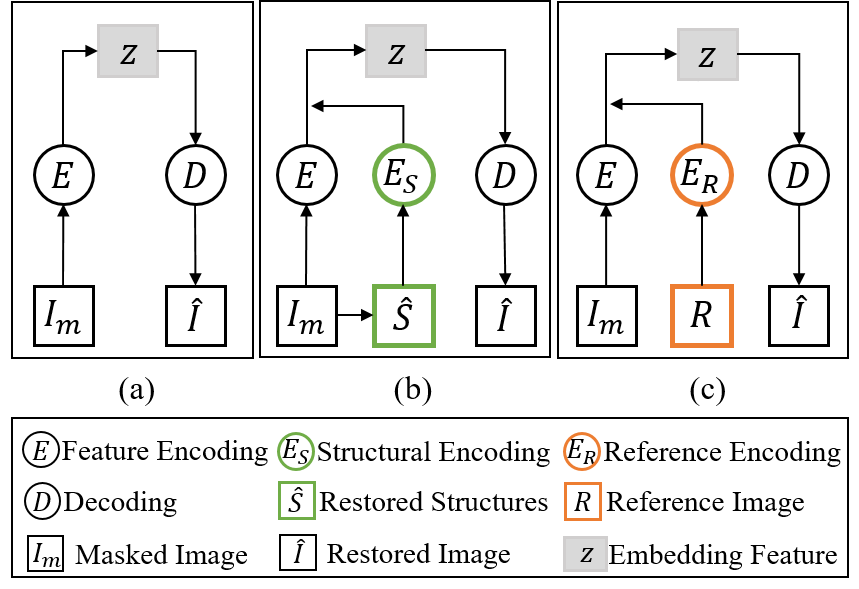}
\caption{Illustration of image inpainting models. (a) Standard context-based inpainting model; (b) Structure-guided inpainting model; (c) The proposed reference-guided inpainting model.}
\label{fig:deepfeature}
\end{figure}

Recently, learning-based inpainting approaches have gained great interest, benefiting from the significant advances in convolutional neural networks (CNNs) and large-scale data collection. Pathak \emph{et al.} first introduced a learning-based inpainting framework, \textit{i.e.}, Context Encoders \citep{ContextEncoders}, which served as the basis for various image inpainting techniques, including neural patch synthesis~\citep{patch}, residual learning~\citep{residual}, partial convolution~\citep{PConv}, and others~\citep{4,MEDFE,liaoeccv} (summarized in Fig.~\ref{fig:deepfeature}(a)). These methods are based on the assumption that the encoded feature of a corrupted image contains sufficient contextual priors for restoring the missing contents ~\citep{liaojstsp}. This assumption is adequate for simple corrupted patterns since the context manifold of such patterns can be reasonably well characterized by a deep model. Nevertheless, when a corrupted region involves multiple semantics, it becomes challenging to uniformly map different semantics onto a single manifold in context-based approaches due to the complicated interactions between the missing pixels and their surroundings~\citep{liaoCVPR,liaoeccv}, which often leads to blurry boundaries and incorrect semantic content. To assist the inpainting of complicated scenes, structural priors such as edges~\citep{EC}, contours~\citep{Contour}, and segmentation maps~\citep{47,liaoeccv} are extracted and restored to guide the inpainting process (\emph{e.g.}, Fig.~\ref{fig:deepfeature}(b)). These methods demonstrate that the structural priors can effectively help improve the completed image quality by providing a preliminary structural composition. However, they still struggle in cases when holes are large or the expected content contains complicated semantic information.

\begin{figure}[tb]
	\centering
		\begin{tabular}{cccc}
			\includegraphics[width=0.245\columnwidth]{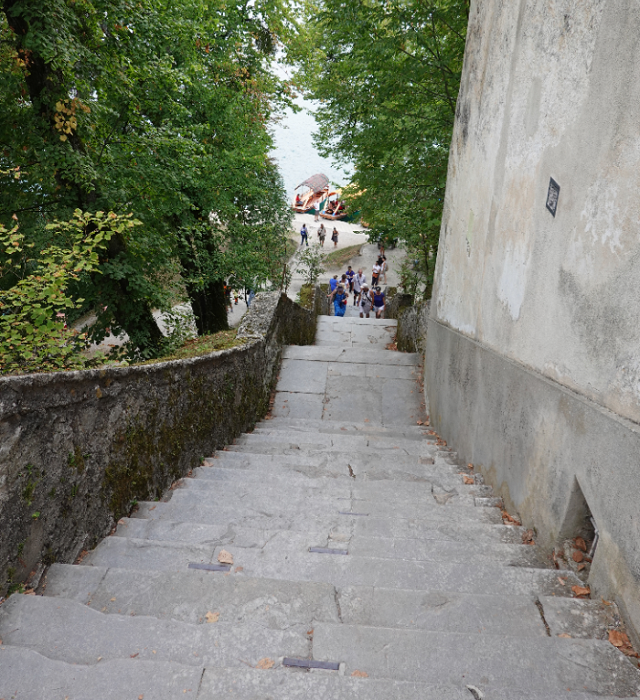} &
			\includegraphics[width=0.245\columnwidth]{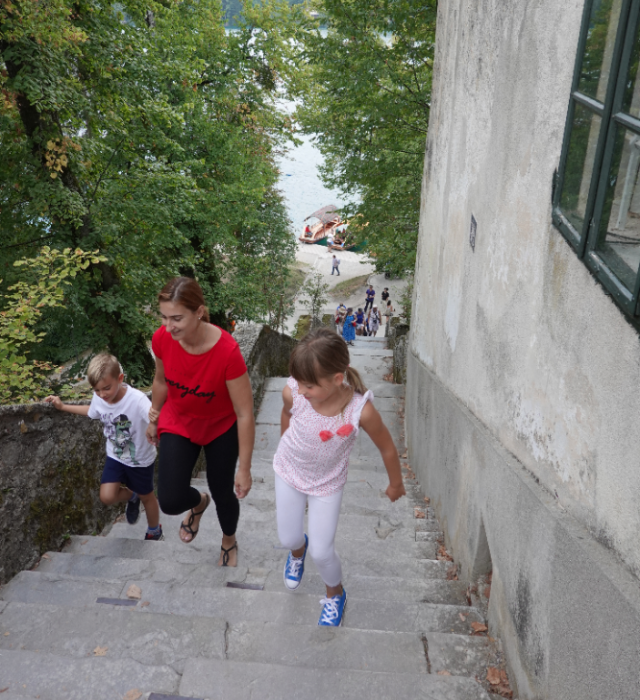} &
			\includegraphics[width=0.245\columnwidth]{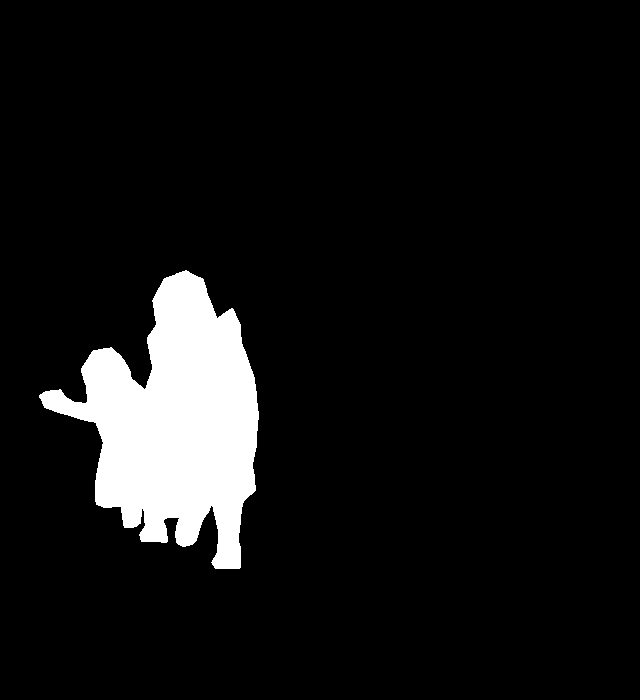} &
			\includegraphics[width=0.245\columnwidth]{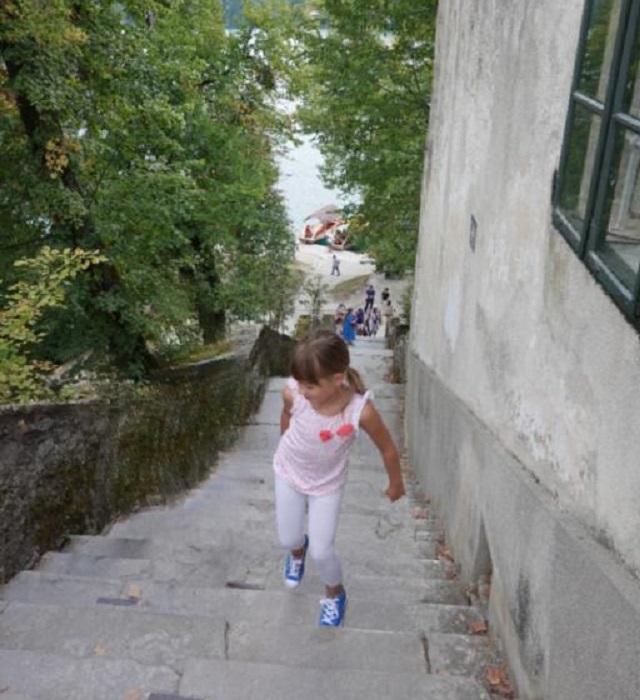}  \\
			\includegraphics[width=0.245\columnwidth]{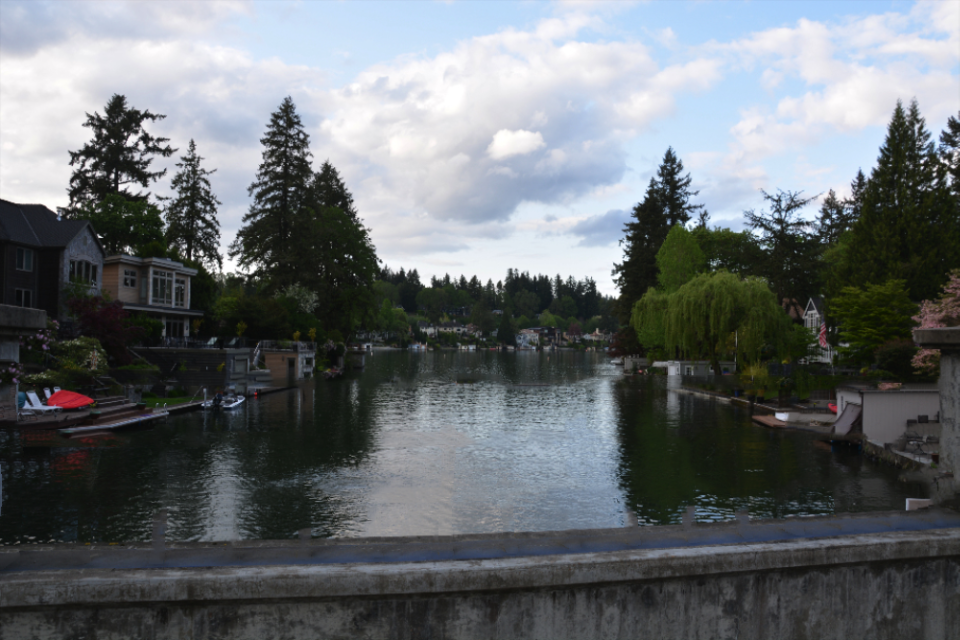} &
			\includegraphics[width=0.245\columnwidth]{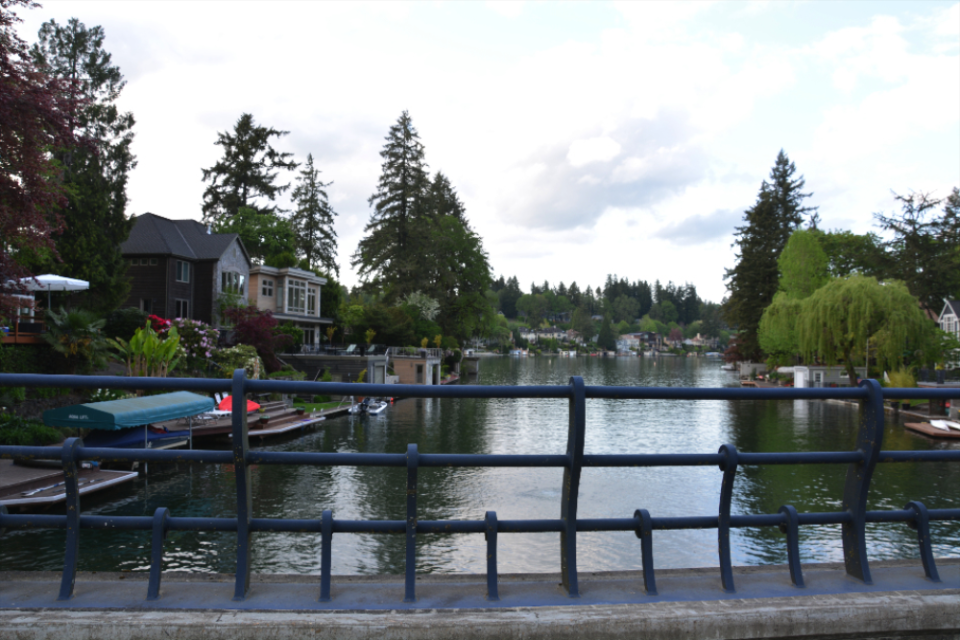} &
			\includegraphics[width=0.245\columnwidth]{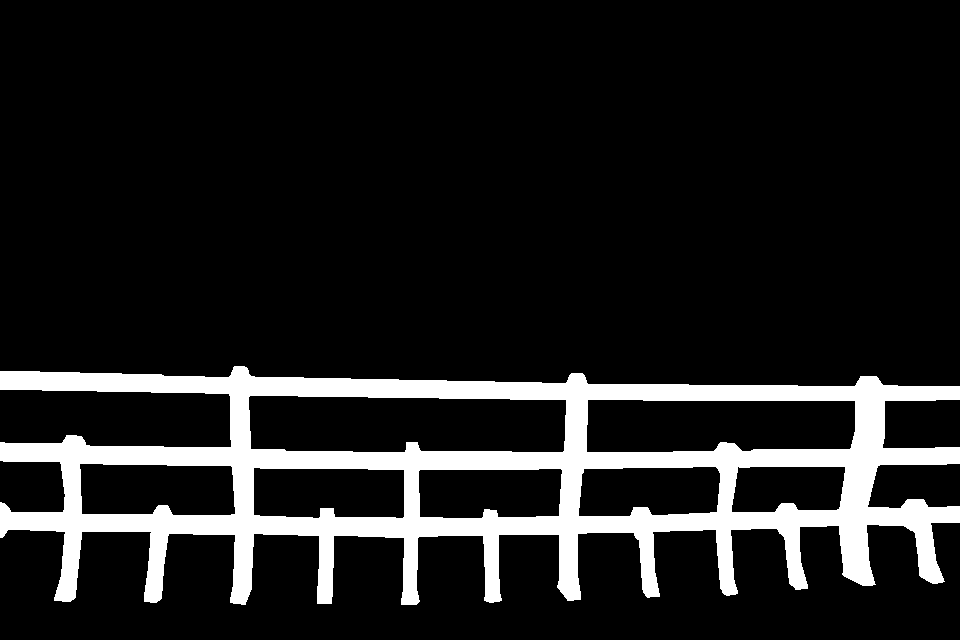} &
			\includegraphics[width=0.245\columnwidth]{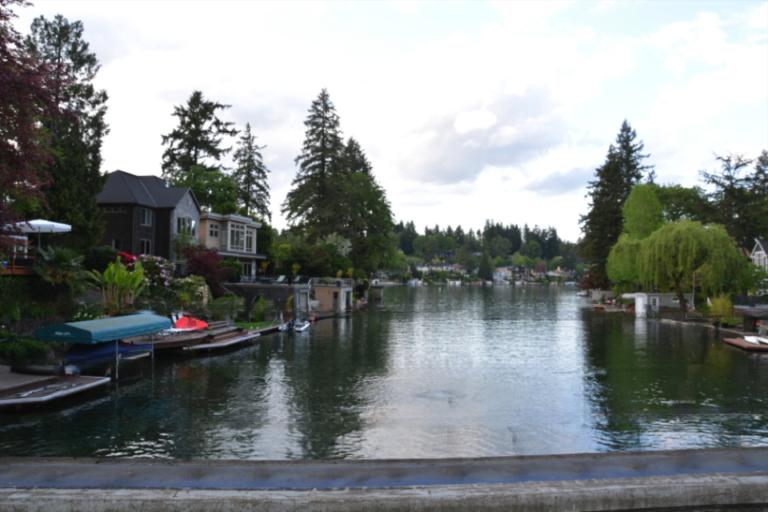}  \\
   			\includegraphics[width=0.245\columnwidth]{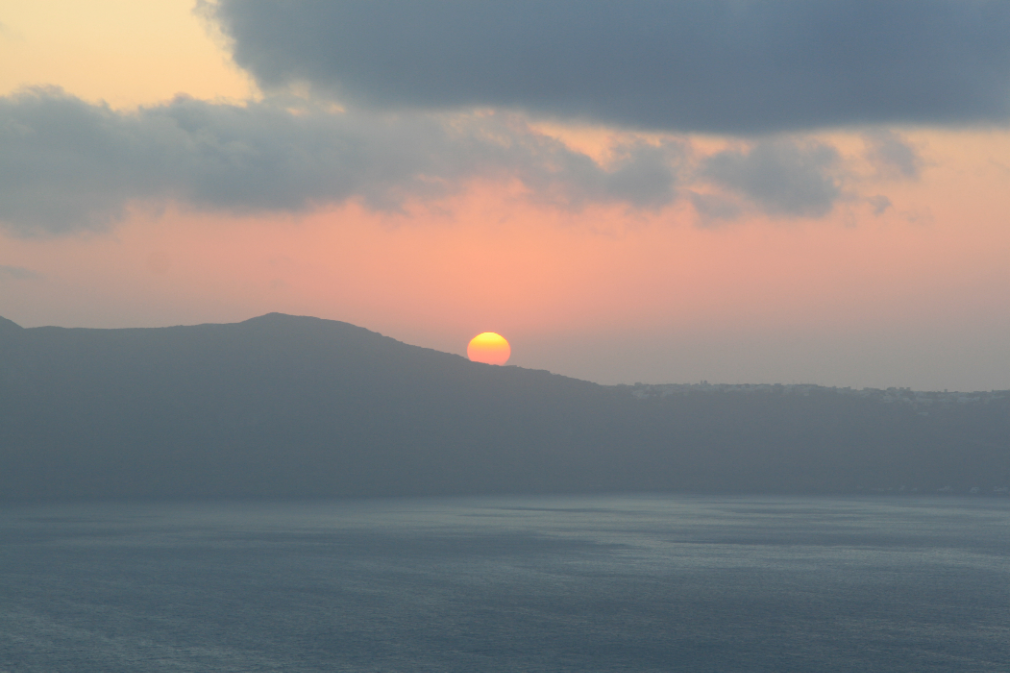} &
			\includegraphics[width=0.245\columnwidth]{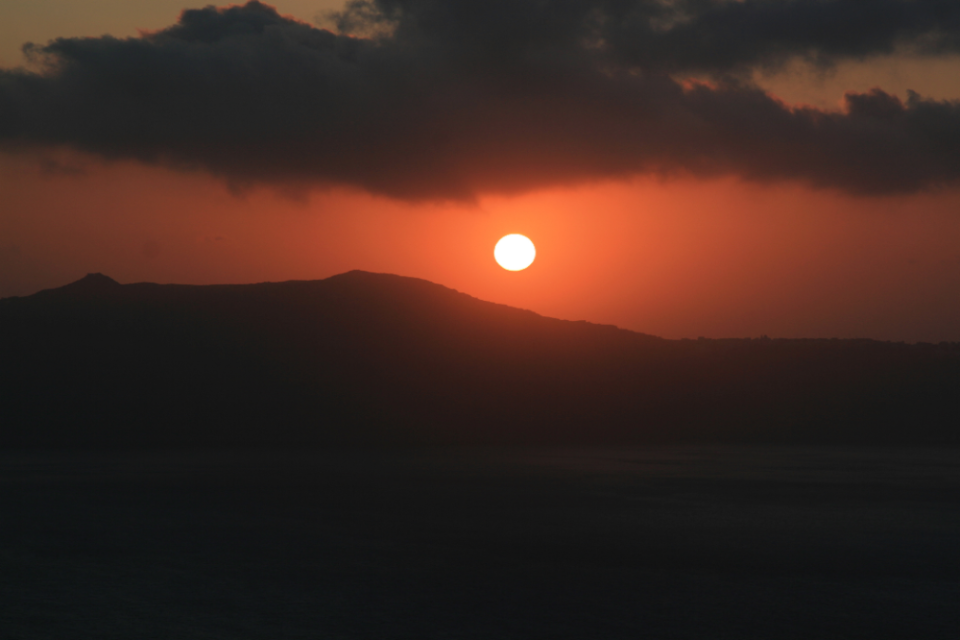} &
			\includegraphics[width=0.245\columnwidth]{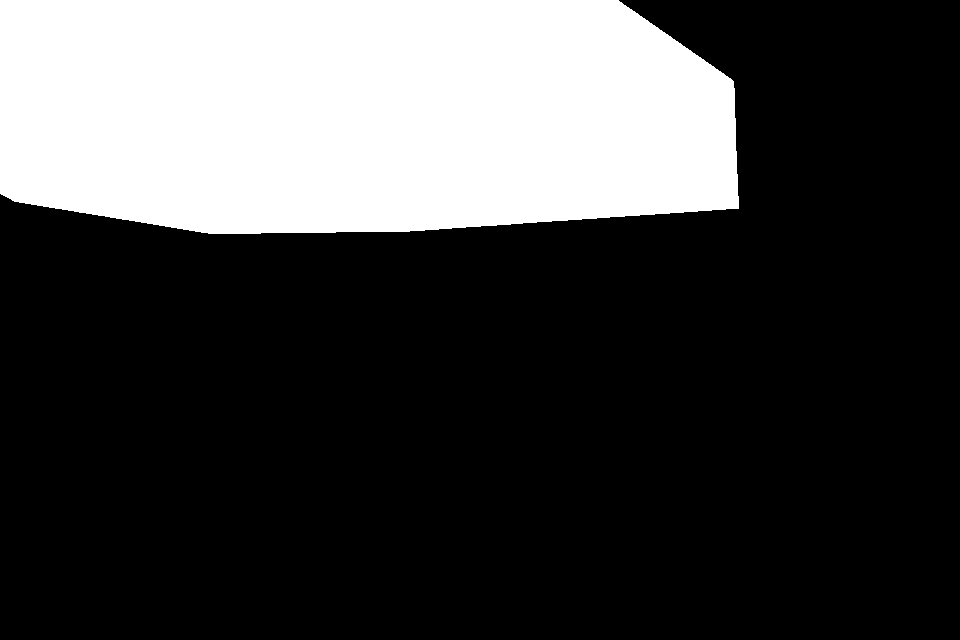} &
			\includegraphics[width=0.245\columnwidth]{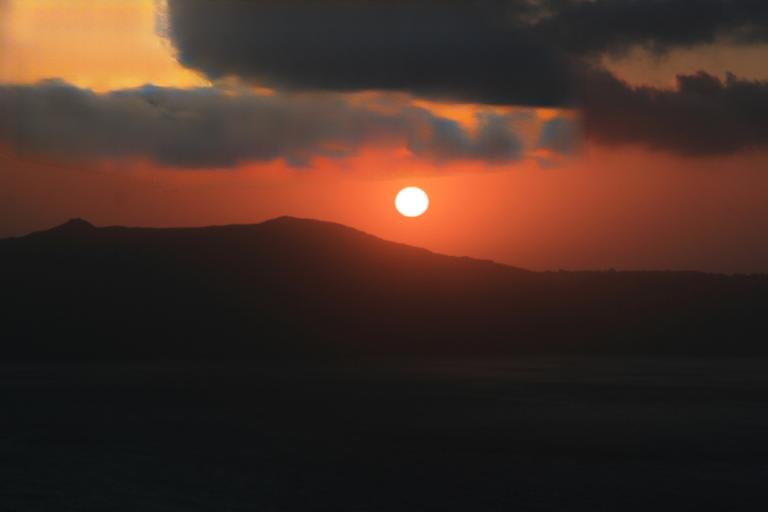}  \\
		\footnotesize{(a) Reference} & \footnotesize{(b) Input} & \footnotesize{(c) Mask} & \footnotesize{(d) \textbf{TransRef}}
	\end{tabular}
\caption{Application scenarios of reference-guided image inpainting, such as object removal (the top and middle rows) and image completion (the bottom row) which all require restoring corrupted scenes to their original state, where the reference images serve as an informative guide to faithfully restore the missing contents.}
	\label{scenario}
 \vspace{-5mm}
\end{figure}

To tackle this ``ill-posed'' problem of image inpainting, we propose to introduce reference images obtained from the web or captured at different viewpoints and dates, which contain similar structures and details but diverse styles and geometries for the filling reference (\emph{e.g.}, Fig.~\ref{fig:deepfeature}(c)). Such a reference would greatly assist the objective of image inpainting, \textit{i.e.}, restoring corrupted scenes to their original state as opposed to producing pluralistic images. For instance, Fig.~\ref{scenario} demonstrates some application scenarios of reference-guided image inpainting, such as object removal (the top and middle rows) and image completion (the bottom row), which all require restoring corrupted scenes to their original state, where the reference images serve as an informative guide to faithfully restore the missing contents. Additionally, reference images have proven to be beneficial for specific vision tasks, such as image super-resolution~\citep{MASA,SRNNT} and image compression~\citep{xiaotmm16,ho2021rr}, as a reference image can compensate for the lost details in a corrupted image by providing rich textures, thereby producing a more realistic and detailed image appearance. In contrast to those existing reference-based vision tasks, reference-guided inpainting is expected to provide complete structural information and detailed textures to define the boundaries between various semantics, leading to more considerable challenges in image alignment and content restoration. 

\tabcolsep=0.5pt

Diffusion models have recently garnered significant attention for their impressive performance in image generation and image restoration tasks \citep{lugmayr2022repaint,yang2023uni,cao2024leftrefill,wang2024frequency,yuan2023efficient}. RSFSG \citep{yuan2023efficient} is proposed to generate controllable remote sensing fake samples consistent with real scenes based on the diffusion model. Uni-paint \citep{yang2023uni} utilizes multi-model guidance to generate diverse inpainting results. FCBlock \citep{wang2024frequency} is proposed to combine frequency information with the diffusion model for the image dehazing tasks. 
However, as for reference-guided image inpainting task which utilizes a reference image shared with similar texture and structure to restore damaged scenes to their original appearance, these diffusion-based methods face several key challenges that impact their overall effectiveness and efficiency.
Firstly, capturing global contextual information effectively is a great challenge for diffusion models. Diffusion models typically rely on local context and iterative refinement, which can be less effective in modeling long-range dependencies and complex interactions within the image compared to transformer-based models. Transformers excel in capturing global context through their self-attention mechanisms, allowing them to integrate information from reference images more accurately and restore damaged scenes more effectively. Secondly, diffusion models inherently involve stochastic processes during the iterative denoising steps, resulting in variability in the generated results, which is not ideal for the consistent restoration required in the reference-guided image inpainting tasks. 
Lastly, the speed of diffusion models is a major concern. The generation process involves multiple iterative steps to progressively refine the image, leading to substantial computational time and increased complexity.

In this paper, we propose a transformer-based reference-guided inpainting framework, named \textbf{TransRef}, built on a multi-scale transformer architecture. Compared with CNNs, the transformer can effectively establish distant pixel correlations and is more suitable for modeling long-range dependency for image inpainting. In each scale of the encoder, the reference information is gradually embedded into the corrupted image. Specifically, a reference patch alignment (Ref-PA) module is proposed to coarsely align and fuse the patch embeddings of the reference image with the masked image, including a spatial alignment process and a style harmonization process to bridge the geometry and style gaps between the reference and masked images. Additionally, to conduct fine-grained feature refinement on the coarsely aligned features, we propose a reference-patch transformer (Ref-PT)  module to refine the fused feature at the mini-patch level through multi-head reference attention, which is then fused with the main feature from the masked image. Finally, the fused features from all scales, representing different levels of features, are concatenated and then decoded into the completed image. 

Besides the proposed framework, to promote this novel task, we construct a publicly accessible reference-guided image inpainting dataset, referred to \textbf{DPED50K}, consisting of a training set of 50,000 input-reference image pairs and a test set of 2,000 pairs. The corresponding reference for each input image is selected using the SIFT~\citep{SIFT} matching algorithm from the \textbf{DPED}~\citep{DPED} dataset, which contains various scenarios, such as street scenes, houses, vehicles, billboards, and plants. We believe that \textbf{DPED50K} can well serve as an open benchmark for performance evaluation and facilitate the development of reference-guided image inpainting tasks.

A preliminary version of this work was presented in our conference version ~\citep{RGTSI}, which presents the basic idea of reference-guided image inpainting. In this paper, we extend the previous work in five aspects: (1) We replace the CNNs with the transformer architecture to fill the image holes, alleviating the limited receptive field size and establishing long-range feature correlations; (2) We upgrade the FAM module in~\citep{RGTSI} to the Ref-PA module by replacing the simple feature adding using a PH block to better harmonize the style difference from the reference image; (3) In addition to coarse alignment in Ref-PA module, we propose a new Ref-PT module to refine the fused feature to make the filled-in features better fit the masked image; (4) We significantly enrich the number of image pairs in our constructed dataset from 10K to 50K so that more realistic scenes can be covered; (5) We conduct additional experiments extensively to evaluate the effectiveness of the proposed \textbf{TransRef}.

The main contributions of our paper are fourfold:
\begin{itemize}
\item We explore the idea of reference-guided image inpainting to complement insufficient information in completing complex scenes, breaking the performance barrier of existing image inpainting methods that cannot guarantee the authenticity of the completed image.% when facing large masks.
\item We propose a transformer-based inpainting network with a multi-scale reference embedding procedure to address the issues of image alignment and content restoration in the presence of missing large regions.   
\item We devise a reference patch alignment (Ref-PA) module to coarsely align the embedded input and reference patches, and a reference patch transformer (Ref-PT) module with the multi-head reference-attention mechanism to further refine the misaligned contents. 
\item We contribute a new dataset \textbf{DPED50K} that consists of a rich set of real-world corrupted and reference image pairs, to facilitate the research and performance evaluation of reference-guided image inpainting tasks.
\end{itemize}

The remainder of this paper is organized as follows. We introduce related work in Sec.~\ref{sec:relatedwork}. The proposed \textbf{TransRef} method is elaborated in Sec.~\ref{sec:method}. In Section~\ref{sec:dataset}, we introduce our newly constructed dataset \textbf{DPED50K}. Experimental settings and extensive experimental results are presented in Sec.~\ref{sec:experiment}. Finally, conclusions are drawn in Sec.~\ref{sec:conclusion}.

\section{Related Work}
\label{sec:relatedwork}
In this section, we briefly review related work in each of the three sub-fields: traditional image inpainting, learning-based image inpainting, and scene priors-guided image restoration.

\subsection{Traditional Image Inpainting}
Traditional image inpainting methods mainly consist of diffusion-based methods~\citep{bertalmio2000image, 1401905} and exemplar-based methods~\citep{PatchMatch,ding2018image}. The basic principle of diffusion-based image inpainting is to model the task with a high-order partial differential equation or variational model to smoothly diffuse the available boundary information from the exterior of the missing region into the interior. Bertalmio \textit{et al.}~\citep{bertalmio2000image} first proposed to spread the Laplacian operator of the adjacent region into the missing region along the isophotes.  Based on this model, subsequent researchers proposed improved methods in terms of diffusion localization information~\citep{li2017localization}, diffusion direction~\citep{Li2016ImageIA}, and model solving method~\citep{sridevi2019image}. However, this method simply utilizes the local image smooth priors along the missing region, neglecting the restoration of image texture. When it deals with the inpainting of large damaged regions, its results are often blurry.

To fill in the large holes, exemplar-based methods borrow patches with a similar appearance. Criminisi \textit{et al.}~\citep{4} proposed to find and paste the most similar patch from the known area to the missing region. PatchMatch~\citep{PatchMatch} provided a multiscale patch-searching strategy that searches for approximate nearest-neighbor matches by random sampling to accelerate the inpainting process. However, their performance remains limited due to the inability to obtain high-level information for context understanding.

\subsection{Learning-based Image Inpainting}
Recent advances in deep learning have yielded impressive results in image inpainting tasks. Pathak \textit{et al.}~\citep{ContextEncoders} incorporated adversarial training~\citep{GAN} to inpainting and utilized an encoder-decoder architecture to fill in missing pixels. Afterward, various variants~\citep{38,PENnet,wang2018image} of the U-Net structure have been developed. In addition, more sophisticated network modules or learning strategies were proposed to generate high-quality images, such as global and local discriminators~\citep{Globally}, contextual attention~\citep{GC}, partial convolution~\citep{PConv} and gated convolution~\citep{GConv}, \emph{etc}. Recent remarkable works have also been proposed to focus on large hole filling~\citep{suvorov2021resolution,lefeicvpr20,PRVS}.

Multiple intermediate hints have been explored extensively to seek powerful guidance to assist in the inpainting of large corrupted regions. Initially, image edges~\citep{EC,Liangedge} were employed as structural guidance to reconstruct missing contents. Later on, Li \emph{et al.}~\citep{Contour} learned to predict foreground contours, which then serve as guidance for restoring the missing region. Structureflow~\citep{structureflow} proposed a structure reconstructor that generates smoothed images as global structural information. Yang \emph{et al.}~\citep{yang2020learning} introduced a structure embedding scheme that can explicitly provide structure preconditions for image restoration. ZITS~\citep{ZITS} used a transformer architecture to incrementally add structural information with masking positional encoding. Other works like SPG-Net~\citep{47} and SGE-Net~\citep{liaoeccv} used semantic segmentation maps, and MMG-Net~\citep{UnbiasedMulti-Modality} used both semantic segmentation maps and edges to learn the structural features efficiently.
%加diffusion？

These methods can restore the corrupted regions with repetitive structures and textures with the support of large-scale data training. However, they continue to suffer from producing results that are semantically implausible or have artifacts in the completed regions, especially when corrupted regions contain distinctive contents. In addition, although pluralistic image inpainting methods~\citep{Pluralistic,UCTGAN,49} were proposed to generate multiple plausible solutions given the missing holes, the completed images cannot be restored to their original state, which is contrary to the key aim of image inpainting.

\subsection{Scene Priors-guided Image Restoration}
Image restoration is the process of restoring an image from a degraded version. As a representative task, single image super-resolution (SISR)~\citep{dong2015image} aims to recover a high-resolution (HR) image from a single low-resolution (LR) image. Different from SISR, reference-based image super-resolution (RefSR)~\citep{MASA,SSEN,CrossNet} was recently proposed to super-resolve input LR images with an extra reference image. By aligning the features from the LR and reference images, RefSR methods can transfer the textures from the reference image to compensate for missing details and improve the restoration results. For instance, SSEN~\citep{SSEN} used deformable convolutions to extract the reference features. Cross-Net~\citep{CrossNet} aligned the LR and reference images based on the estimated flows between the two images. C2-Matching~\citep{C2} introduced contrastive learning and knowledge distillation into the patch-matching stage, bridging the transformation gap between LR and reference images. 

Recent research~\citep{tranfill,GeoFill,9840396,RGTSI,yang2023uni,cao2024leftrefill} has introduced a reference-based strategy to image inpainting tasks. Transfill~\citep{tranfill} proposed a multi-homography transformed fusion approach to fill the hole by referencing an additional image captured simultaneously but from different viewpoints. DFM ~\citep{9840396} presented a two-step warping technique to recover the corrupted image, which is robust to style-inconsistent reference images by transferring structural patterns. RGTSI~\citep{RGTSI} introduces a strategy to fuse texture and structure from image pairs with various styles and geometries. Uni-paint~\citep{yang2023uni} utilize text, strokes, or an exemplar image to fine-tune the diffusion model, thus generating diverse inpainting results. LeftRefill~\citep{cao2024leftrefill} based on a generalized text-to-image diffusion model employs either a single image or multiple viewpoints for image inpainting or novel view synthesis.
Despite the impressive performance of the recent methods, their applications are limited and lack sufficient priors in restoring general scenes especially when the missing part contains complex semantics. To mitigate these issues, we propose a novel reference-guided model and an advanced feature alignment module to finely align and fuse the feature of the input and reference images.

\section{Proposed Method}
\label{sec:method}
In this section, we describe the proposed \textbf{TransRef}. We first introduce the problem formulation of the reference-guided image inpainting task. Next, we provide an overview of our framework, especially the reference patch alignment and the reference patch transformer, followed by the transformer decoder and the associated objective functions. 

\subsection{Problem Formulation}
Let $I$ denote the original image, $M$ a binary mask with 0 and 1 indicating the known and missing pixels, respectively, and $I_{m}$ the corrupted image formulated as $I_{m} =I\odot(\textbf{1}-M)$, where \textbf{1} is a matrix with all one entries and $\odot$ denotes element-wise multiplication. The goal of image inpainting is to produce a synthetic image $\hat{I}$ reasonably close to  $I$ given the corrupted image and the mask. Typical learning-based image inpainting frameworks employ an encoder-decoder network architecture consisting of an encoder $\mathrm{Enc}$ and a decoder $\mathrm{Dec}$, which is  formulated as
\begin{equation}
    \begin{split}
        & z = \mathrm{Enc}(I_{m},M;\theta_\text{enc}),\\
        & \hat{I}= \mathrm{Dec}(z;\theta_\text{dec})
    \end{split} 
\end{equation}
where $z$ is the latent feature vector encoded by $\mathrm{Enc}$. $\theta_\text{enc}$ and $\theta_\text{dec}$ are the learnable parameters of the encoder and decoder, respectively.

Rather than only inferring the image conditioned on $I_{m}$ and $M$, reference-guided image inpainting introduces a reference image $I_\text{ref}$ as the inpainting guidance, aiming to synthesize an image that recovers the original scene content while reasonably being harmonized with the non-missing content. Specifically, the reference-guided image inpainting problem can be formulated as follows:
\begin{equation}
    \begin{split}
        & z =  \mathrm{Enc}(I_{m},M,I_\text{ref};\theta_\text{enc}),\\
        & \hat{I}= \mathrm{Dec}(z;\theta_\text{dec})
    \end{split} 
\end{equation}
where the introduced reference image $I_\text{ref}$ contains similar content and texture to $I$ and is capable of providing sufficient priors about the scene and compensating for the loss of texture details and structures. 

\begin{figure*}[t]
    \centering    \includegraphics[width=\textwidth]{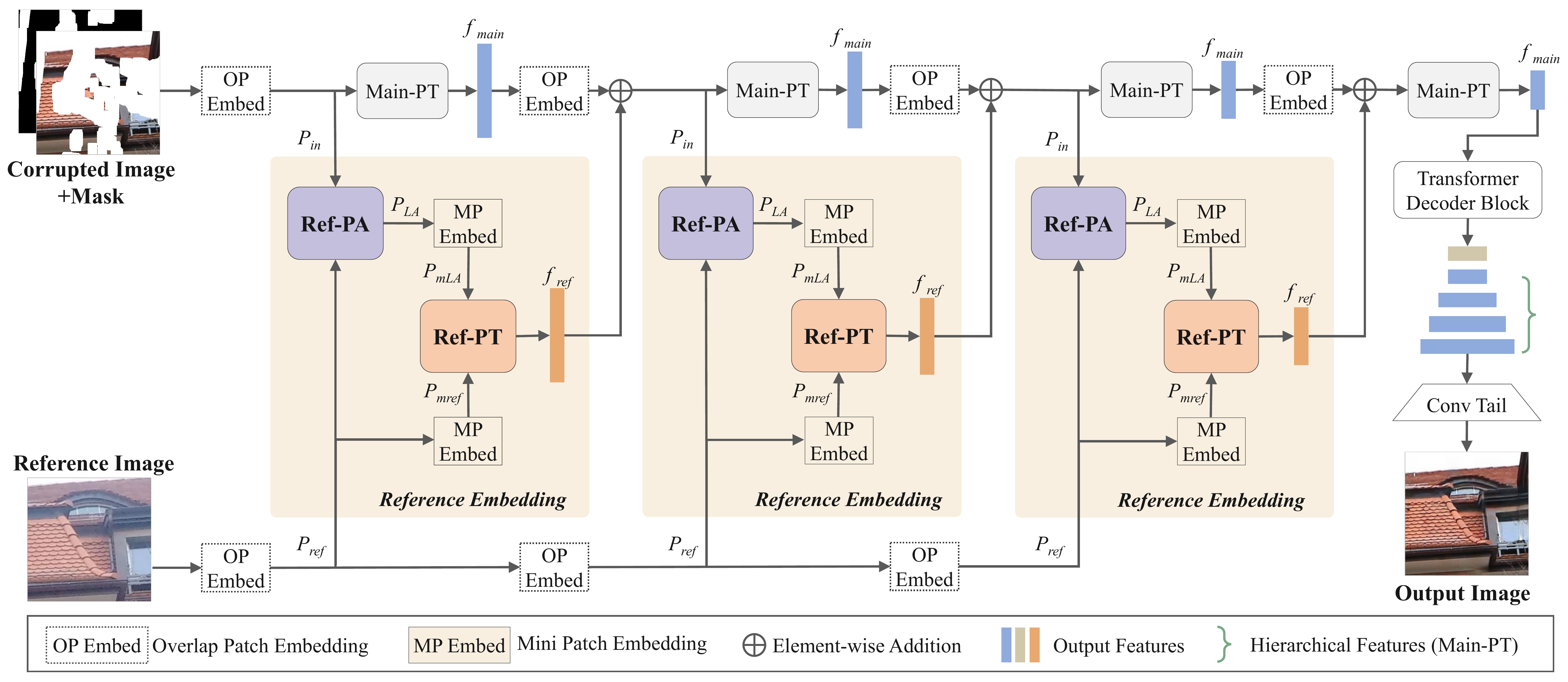}
    \caption{Overview of the proposed \textbf{TransRef}. The first row composed of the overlap patch embedding and the Main-PT modules form the basic hierarchical inpainting framework. The reference guidance is conducted through the reference embedding procedure at each scale by the Ref-PA and Ref-PT modules. In the last, the hierarchical features from the Main-PT module and the decoder features from the transformer decoder block are fed into a convolution tail to generate the completed image.}
    \label{fig:pipeline}
\end{figure*}

\subsection{Framework Overview}

To involve the reference image in the inpainting process, we propose a reference-guided inpainting framework, named \textbf{TransRef}, built on a hierarchical transformer architecture. As depicted in Fig.~\ref{fig:pipeline}, in each scale-branch of the encoder, a main-patch transformer (Main-PT) module~\citep{transweather} is adopted to find the long-range dependencies of the corrupted-image features. Besides, the proposed reference embedding procedure aligns the reference-image features with the corrupted-image features, which are then fused with the features from the Main-PT module to derive the features for filling the missing content. Additionally, the multi-scale framework guarantee the full utilization of reference information of various granularities, including structures and textures, by combining the features from fine to coarse granularity.

%重点介绍我们的contributions in the reference embedding procedure
In general, feature alignment is more challenging in image inpainting than in existing reference-based restoration tasks for the following reasons: 1) sorely relying on the information surrounding the holes leads to distorted image features, preventing accurate feature alignment between the corrupted and reference images; 2) the structures and semantic compositions of holes should be transferred from the reference image and be harmonized with the existing surroundings. 
To address the problem of aligning corresponding image content with large holes, the reference embedding procedure is featured with two new modules to align the features of the reference image with that of the corrupted image in a local-to-global manner. 
At the local level, the reference patch alignment (Ref-PA) module is proposed to coarsely align the patch embedded from the reference image towards the input image by deformable convolution~\citep{Deform} and then to transfer the style of reference-image features towards that of the corrupted-image features to achieve style harmonization. At the global level, since the local alignment on distorted features may result in inaccurate alignment in some local areas, we propose a reference patch transformer (Ref-PT) module with a multi-head reference-attention mechanism to further refine misaligned contents by utilizing the small-patch-size features from the mini-patch embedding to extract more details.

Specifically, the inpainting process of \textbf{TransRef} can be summarized as follows. In each scale of the reference embedding, Ref-PA first aligns the reference patches locally with the input patches and fuses the reference-image features with the corrupted-image features for coarse filling. Then, Ref-PT globally refines misaligned contents by calculating the multi-head reference attention between the reference patches and the aligned patches after mini-patch embedding. The features from Ref-PT are added to those from Main-PT, 
and then fed into the next-scale encoding stage. When the reference patch alignments of the three scales
%, \textit{i.e.}, $\frac{1}{4}$, $\frac{1}{8}$, and $\frac{1}{16}$, 
are completed, the reference-image information has been adequately aligned with the corrupted-image features for restoring the corrupted areas. Lastly, the multi-scale embeddings from the Main-PT modules are decoded by the transformer decoder, followed by a convolutional tail to generate the completed image.

\begin{figure}[t]
    \centering
    \includegraphics[width=0.85\columnwidth]{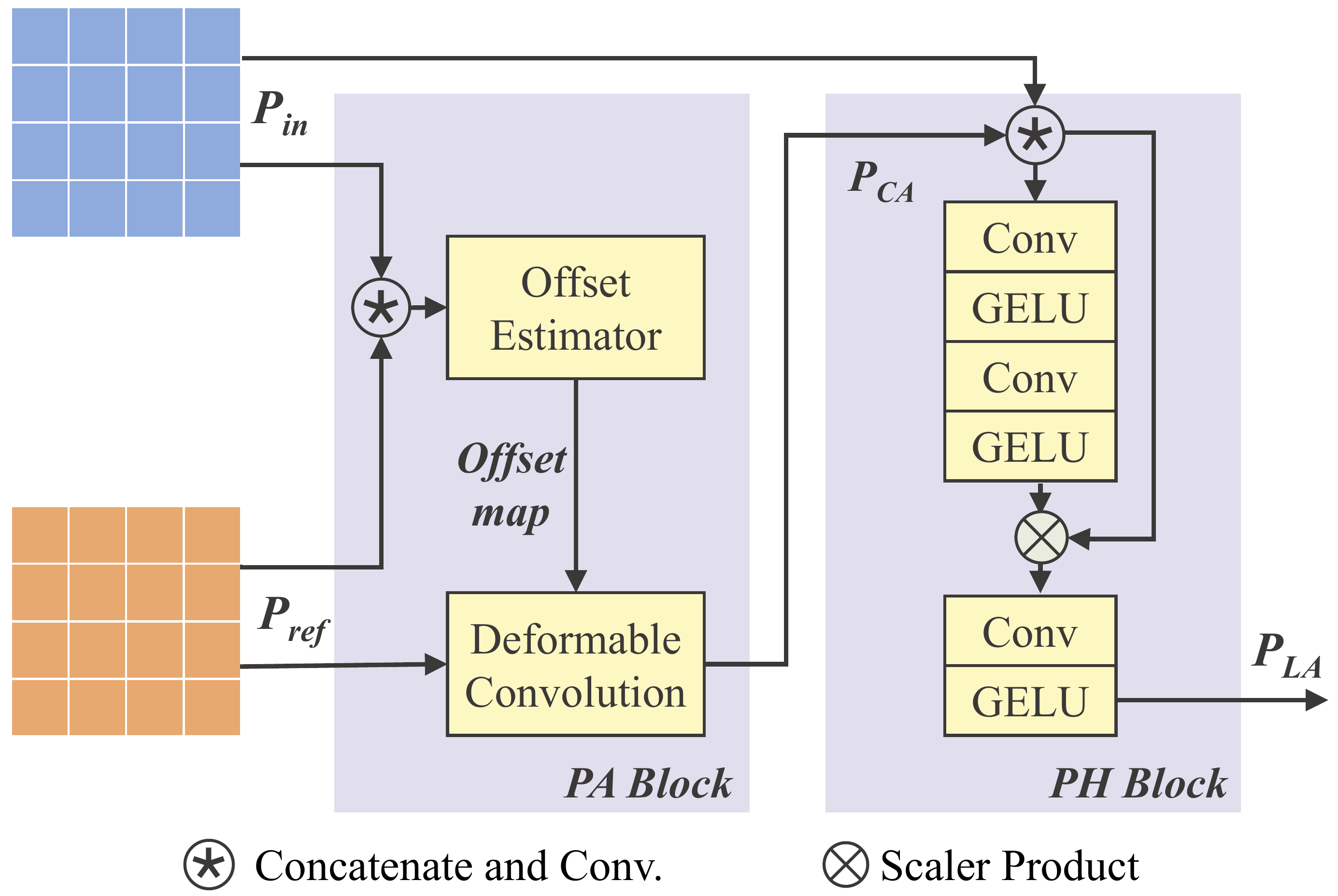}
        \caption{Illustration of the Ref-PA module. The Ref-PA consists of the PA block for patch alignment and the PH block for patch harmonization.}
    \label{fig: PAM}
\end{figure}

\subsection{Reference Patch Alignment (Ref-PA)}

Ref-PA is designed to locally align the reference patches with the input image to produce coarsely aligned and transferred patch features. It receives the patch features from both the input and reference images after overlapped patch embedding~\citep{segformer}, which can extract features with better local continuity to benefit the alignment of patches around the masked areas than non-overlapped patch embedding. Ref-PA consists of a patch alignment (PA) block and a patch harmonization (PH) block to learn the offset map for structure alignment and to harmonize the texture styles for the inferred feature, respectively. The procedure of Ref-PA can be formulated as follows: 
 \begin{equation}
 \begin{aligned}
     P^i_\text{LA} = \mathrm{RefPA}(P^i_\text{in}, P^i_\text{ref}),\\
 \end{aligned}
 \end{equation}
where $\mathrm{RefPA}$ denotes the Ref-PA module, $P_\text{in}$ and $P_\text{ref}$ are the patch information from the input and reference images, respectively, and $P_\text{LA}$ is the locally aligned feature. $i$ represents the $i$-th scale in the encoder, which is omitted when the context is clear in the remainder. 

After performing $\mathrm{RefPA}$, the guidance information extracted from the reference image can be coarsely aligned to help fill the holes in the input image.

%XJ：这个part组织的逻辑：
%1）一小段话总述：做什么事情，解决什么问题，得到什么结果
%2）input的是什么：overlap patch embedding的内容及原因
%3）local align的思路是什么？具体利用什么信息，用什么方法做alignment（具体的PAM结构，名字根据overview调整一下），如何克服空洞的问题
%4）align之后，是否做了融合，输出了什么feature

\subsubsection{Patch Alignment Block}
In this paper, we follow ~\citep{RGTSI} to use the deformable convolution~\citep{Deform}, which models deformation by learning an offset map to achieve local alignment. This offset enables the model to focus on the target region when extracting features by shifting its convolution kernels, hence tackling the problem of insufficient information in the masked areas. Formally, the deformable convolution operation is defined as follows: 
\begin{equation}
    Y(p_0) = \sum_{n=1}^{N} \omega_n \cdot X(p+p_n+\Delta p_n),
\end{equation}
where $X$ denotes the input, $Y$ denotes the output, and $n$ and $N$ are the index and the total number of kernel weights, respectively. $\omega_n$, $p$, $p_n$, and $\Delta p_n$ are the $n$-th kernel weight, the central index, the $n$-th fixed offset and the learnable offset for the $n$-th position, respectively. 

As illustrated in Fig.~\ref{fig: PAM}, the feature embedding from the input $P_\text{in}$ and the reference $P_\text{ref}$ are concatenated and then passed to the dynamic offset estimator~\citep{SSEN} to capture similarities from near to far distances and produce an offset map between the input and reference images. Guided by the offset map, the deformable convolution process~\citep{Deform} aligns the reference patches and generates the coarsely aligned feature $P_\text{CA}$. 

\subsubsection{Patch Harmonization Block}
When the information of the reference image is transferred to the holes of the input image, it should look harmonious with the surroundings of the holes. Therefore, inspired by SELayer ~\citep{selayer}, which can adaptively recalibrate channel-wise feature responses by explicitly modeling interdependencies between channels, the proposed PH block is focused on the channel attention map to harmonize the styles of the input and reference patch features by transferring the locally aligned reference features to the input features.

As depicted in Fig.~\ref{fig: PAM}, we concatenate the input patch features and the locally aligned reference features and perform a linear transformation through two convolutional layers with a kernel size of $1\times1$. It is worth noting that the PH block is focused on the local attention map in the input patches between two channels to preserve the information of both channels. 
Gaussian Error Linear Unit (GELU)~\citep{gaussian} is used as the activation function following each convolutional layer, as it avoids the problem of vanishing gradient and also performs better in transformer~\citep{bert}. After the fully connected layer, we employ a $1\times1$ convolution kernel together with GELU again to reduce the feature dimensions and produce the final harmonized locally-aligned feature $P_\text{LA}$.

\subsection{Reference Patch Transformer (Ref-PT)}

To obtain a more precise feature alignment, we propose a Ref-PT module that performs global correction of misaligned content after the coarse alignment of Ref-PA. In contrast to Ref-PA, which uses a deformable convolution guided by the computed offset map to focus on the local alignment of the target region, Ref-PT employs an elaborate multi-head reference-attention mechanism to focus on the alignment of global contents. 

\begin{figure}[tb]
	\centering
		\begin{tabular}{c}
		\includegraphics[width=0.9\columnwidth]{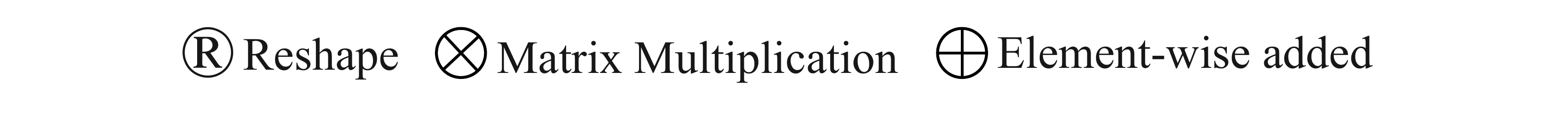}  \\		\includegraphics[width=0.9\columnwidth]{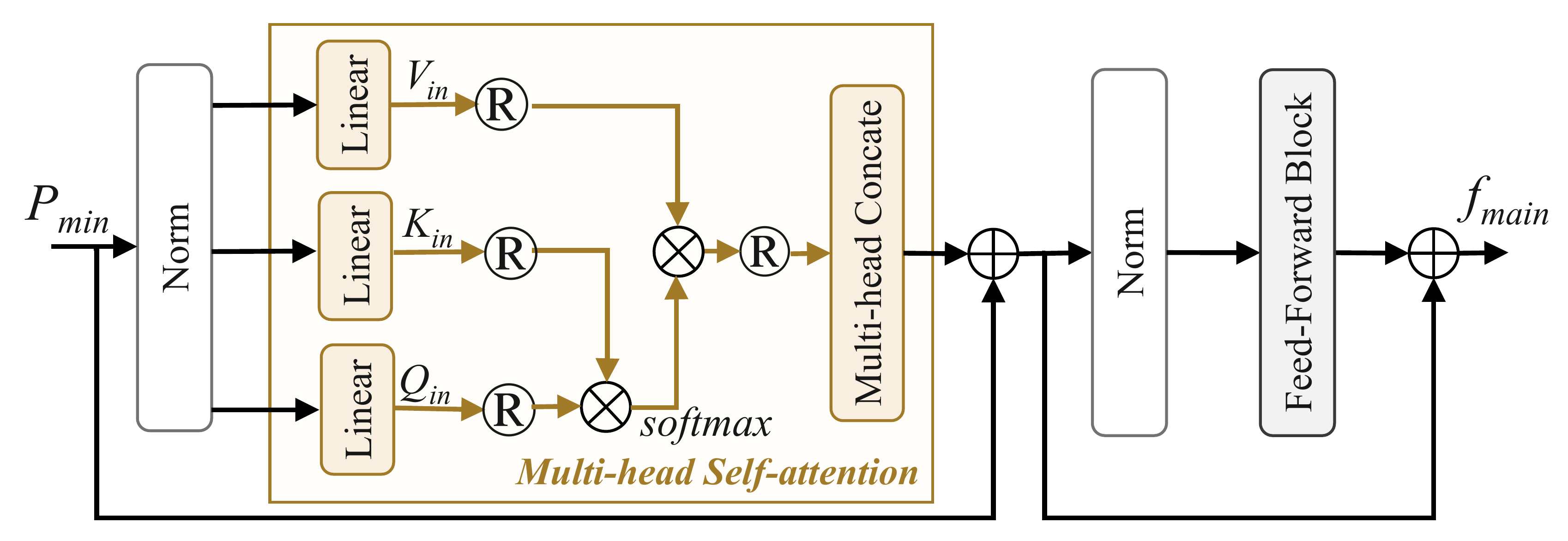}  \\	
			\footnotesize{(a) Main-Patch Transformer Module} \\
			\includegraphics[width=0.9\columnwidth]{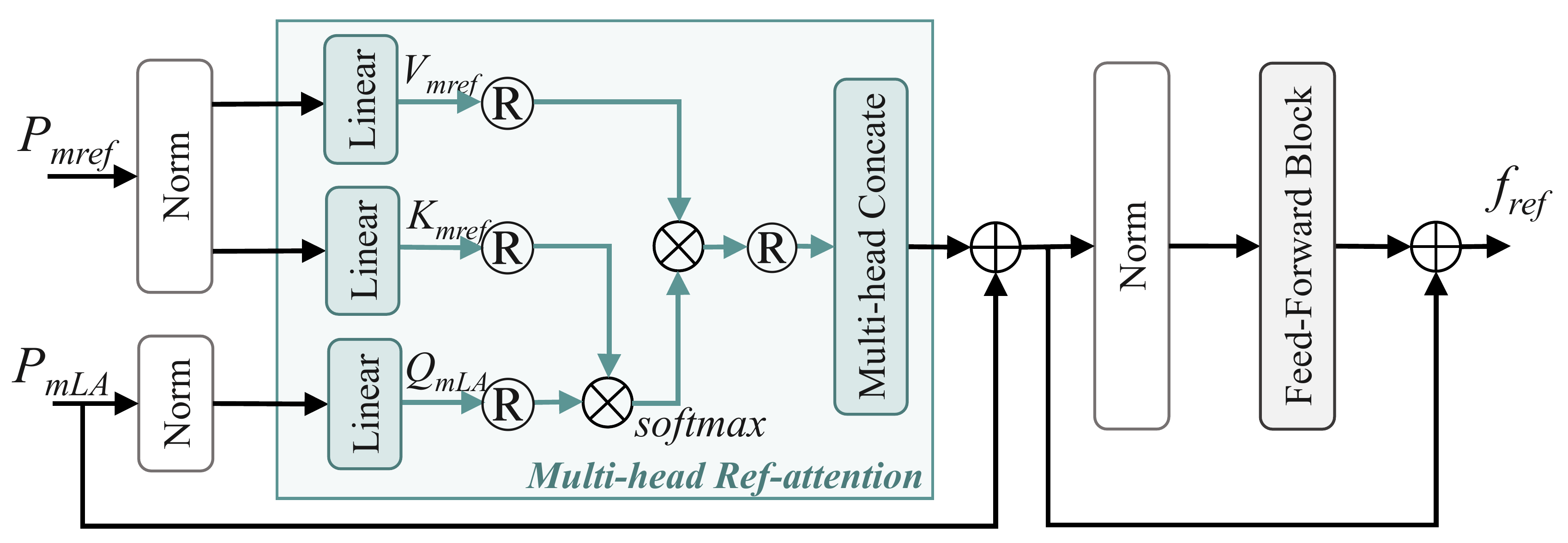}  \\
			\footnotesize{(b) Reference-Patch Transformer Module}  \\
	\end{tabular}
   \caption{Illustration of (a) the Main-Patch Transformer (Main-PT) Module and (b) the Reference-Patch Transformer (Ref-PT) Module.}

\label{fig:TB}
\end{figure}

Specifically, we first perform mini-patch embedding on the input features of Ref-PT, which adopts a smaller patch size than overlapped patch embedding. The mini-patch embedding reduces the spatial dimension of the reference patches $P_\text{ref}$ and the locally aligned patches $P_\text{LA}$ by half to extract fine details for refining the misaligned content. 
Ref-PT takes the mini-embedded features $P_\text{mLA}$ and $P_\text{mref}$ as inputs to integrate the reference features into plausible areas. 

The process can be formulated as follows: 
\begin{equation}
    P_\text{mLA}  = \mathrm{MPE}(P_\text{LA}),
\end{equation}
\begin{equation}
    P_\text{mref}  = \mathrm{MPE}(P_\text{ref}),
\end{equation}
\begin{equation}
\begin{aligned}
    f_\text{ref}  &= \mathrm{RefPT}(P_\text{mref},P_\text{mLA})\\
    &= \mathrm{FFB}(\mathrm{RA}(P_\text{mref},P_\text{mLA}) + P_\text{mLA}),
\end{aligned}
\end{equation}
% \begin{equation}
% f_{out}^i  = f_{main}^i + f_{ref}^i,
% \end{equation}
where $\mathrm{MPE}$ denotes the mini-patch embedding, $\mathrm{RefPT}$ denotes the Ref-PT module. $\mathrm{RA}$ represents multi-head reference-attention, and $\mathrm{FFB}$ denotes the feedforward block. $f_\text{ref}$ is the output feature of  Ref-PT. 

To better illustrate the design of Ref-PT, we compare it with the Main-PT module in the general transformer framework, as depicted in Fig.~\ref{fig:TB}. 
% Given three feature vectors (\textit{query}, \textit{key}, \textit{value}), the attention mechanism first conducts scaled dot-product between \textit{query} ($\mathbf{Q}$) and \textit{key} ($\mathbf{K}$) to produce the similarity map between them. Then a matrix multiplication is performed between the similarity map and \textit{value} ($\mathbf{V}$) to obtain the final result. Throughout this process, the heads of $\mathbf{Q}$, $\mathbf{K}$, and $\mathbf{V}$ have the same dimensions and are calculated by
% \begin{equation}
%     \mathrm{Attn}(\mathbf{Q},\mathbf{K},\mathbf{V}) = \mathrm{softmax}(\dfrac{\mathbf{Q}\mathbf{K}^T}{\sqrt{d}})\mathbf{V},
% \end{equation}
% where $d$ is the channel number of the feature vectors. 
In the Main-PT module, the multi-head self-attention $SA$ is calculated as follows:
\begin{equation}
    \mathrm{SA}(P_\text{in}) = \mathrm{softmax}(\dfrac{\mathbf{Q}_\text{in}\mathbf{K}_\text{in}^T}{\sqrt{d}})\mathbf{V}_\text{in},
    \label{mainPT}
\end{equation}
where the heads of $\mathbf{Q}_\text{in}$, $\mathbf{K}_\text{in}$ and $\mathbf{V}_\text{in}$ are obtained from $P_\text{in}$ through linear projection. 
Unlike the self-attention mechanism in the Main-PT module, where $\mathbf{Q}$, $\mathbf{K}$, and $\mathbf{V}$ are taken from the same input, we modify the multi-head reference-attention in Ref-PT to better integrate the reference information. Specifically, $\mathbf{Q}_\text{mLA}$ is linearly projected from the locally-aligned patches $P_\text{mLA}$, while $\mathbf{K}_\text{mref}$ and $\mathbf{V}_\text{mref}$ are taken from the reference patches $P_\text{mref}$, which preserve more complete information without disruption. The multi-head reference-attention $\mathrm{RA}$ is formulated as
\begin{equation}
    \mathrm{RA}(P_\text{mref},P_\text{mLA}) = \mathrm{softmax}(\dfrac{\mathbf{Q}_\text{mLA}\mathbf{K}^T_\text{mref}}{\sqrt{d}})\mathbf{V}_\text{mref}.
\end{equation}
The reason for setting $\mathbf{Q}$, $\mathbf{K}$, and $\mathbf{V}$ in this way is that the information of the aligned patches after coarse alignment is still incomplete. It can serve as the information to be queried ($\mathbf{Q}_\text{mLA}$), and we should find the expected value ($\mathbf{V}_\text{mref}$) through the keys ($\mathbf{K}_\text{mref}$). In this procedure, the reference patches act as both $\mathbf{K}$ and $\mathbf{V}$. Additionally, since $\mathbf{Q}$, $\mathbf{K}$, and $\mathbf{V}$ are derived by a linear projection of all patches, we believe that the multi-head reference-attention mechanism can focus well on the global information of both the aligned patches and the reference patches.

\subsection{Transformer Decoder}
According to the basic design of transformer-based encoder-decoder, \textbf{TransRef} includes transformer decoder blocks in the decoding stage to account for the consistency of the patch information into feature information. Instead of using all transformer decoder blocks, a convolution tail is also adopted mainly due to two reasons. First, it is designed for fast upsampling to reduce computational complexity and memory cost. Second, we empirically find that this design is superior to the linear projection head used in ViT~\citep{vit} to refine high-frequency details~\citep{li2022mat}, leaning upon the capability and efficiency of CNNs in local texture refinement.

% \subsubsection{Transformer Decoder Block}
% The decoding stage of \textbf{TransRef} operates at a single stage with one transformer decoder block, the architecture of which is similar to that of the Main-PT module. It receives the patch features obtained from the inner Main-PT module and generates the decoded features, which are harmonized with the features extracted across the Main-PT modules in the transformer encoder at each scale. All of these features are concatenated and forwarded to the convolution tail to reconstruct the completed image.

% \subsubsection{Convolution Tail}
% The convolution tail takes in the hierarchical features from the encoding stage and the decoder features from the decoding stage. It comprises four residual blocks to generate the completed image. To restore the original resolution of the input image, we apply two upsampling convolutional layers before the first residual block and one upsampling convolutional layer before each of the other three residual blocks. Skip connections are also used across scales of the convolution tail.
%from the transformer encoder. 

\subsection{Objective Functions}
We design appropriate supervised loss terms for measuring the differences between the generated images and the original images during the training period, including pixel reconstruction loss, perceptual loss, and style loss. 

\subsubsection{Pixel Reconstruction Loss} We use the $\mathcal{L}_1$ loss to encourage per-pixel reconstruction accuracy, formulated as:
\begin{equation}
    \mathcal{L}_{1} = \parallel I_\text{out} - I_\text{GT}\parallel,
\end{equation}
where $I_\text{out}$ denotes the output of \textbf{Transref}, and $I_\text{GT}$ is the corresponding ground-truth.

\subsubsection{Perceptual Loss} Since the reconstruction loss struggles to capture high-level semantics and simulate the human perception of image quality, we use the perceptual loss $\mathcal{L}_{p}$ to penalize the discrepancy between the extracted high-level features:
\begin{equation}
\mathcal{L}_{p} = \mathbb{E} _j\lbrack\sum_{i} \frac{1}{N_i} \parallel \phi_i(I_\text{out}) - \phi_i(I_\text{GT})\parallel_1\rbrack,
\end{equation}
where $\phi_i$ is the activation map of the $i$-th pooling layer of the pretrained VGG-16 network~\citep{SimonyanVGG}. In our implementation, $\phi_i$ corresponds to layered features $relu1\_1$, $relu2\_1$, $relu3\_1$, $relu4\_1$, and $relu5\_1$ in VGG-16 pre-trained on ImageNet.

\subsubsection{Style Loss} To ensure style consistency, we further include the style loss. Given feature maps of size $C_j \times H_j \times W_j$, we compute the style loss as follows: 
\begin{equation}
\mathcal{L}_{s} = \mathbb{E}_j[ \parallel G^\phi_j(I_\text{out}) - G^\phi_j(I_\text{GT})\parallel_1],
\end{equation}
where $G^\phi_j$ is a $C_j \times C_j$  Gram matrix constructed from the activation maps $\phi_j$, which are the same as those used in the perceptual loss. 

In summary, the joint loss is written as
\begin{equation}
\begin{aligned}
\mathcal{L}_\text{joint} = \lambda_{l_1} \mathcal{L}_{1} + \lambda_{p} \mathcal{L}_{p} + \lambda_{s} \mathcal{L}_{s},
\end{aligned}
\end{equation}
where  $\lambda_{l_1}$, $\lambda_{p}$, and $\lambda_{s}$ are the trade-off parameters.

\section{New Dataset for Reference-Guided Image Inpainting}
\label{sec:dataset}

% \subsection{Dataset Construction}

% For each input-reference image pair of size $W \times H$, we first divide them into four equal-size non-overlapping $W/2 \times H/2$ sub-images as well as crop one $W/2 \times H/2$  sub-image from their centers, resulting in five sub-images of the same size. This step aims to reduce the computation for subsequent SIFT feature matching while preserving as much as possible the important details and discriminant features of the two images in a pair. 
% Next, two SIFT objects are created to calculate the key points and their feature descriptors of the target and reference sub-images. With the KNN algorithm, we match the key points and filter the matched feature descriptors to determine the eligible key points by calculating the Euclidean distance. 
% Finally, the input image and the reference image are obtained by cropping the center of the eligible key points in the source and target image patches, with a size of $256 \times 256$, respectively. 
% In this way, we collect a dataset named \textbf{DPED50K} of 52K paired patches, including a training set of 50K pairs and a test set of 2K pairs. 

\subsection{Dataset Overview}
For reference-guided image inpainting, the similarity between the image and its reference image is of great significance to the inpainting results. In this work, we construct a new dataset, namely \textbf{DPED50K}, expended from \textbf{DPED10K} dataset in~\citep{RGTSI}, which consists of real-world photos captured by three different mobile phones and one high-end reflex camera. These photos were taken during the daytime at a wide variety of places and under various illumination and weather conditions. We crop the matched patches from the target image and its reference based on SIFT~\citep{SIFT} feature matching, forming the input-reference pairs in \textbf{DPED50K}. Regarding the test mask patterns, the irregular mask dataset for the evaluation is obtained from~\citep{PConv}. Following the rules of \citep{PConv}, we set six different mask ratios to reorganize the masks, which increase from 0\% to 60\% at a rate of 10\%. Each ratio has 2,000 masks, half of which contain damaged image boundaries and half of which do not. 

\textbf{DPED50K} is a large, accessible, and diverse dataset, consisting of {street scenes}, {houses}, {vehicles}, {billboards}, {plants}, and other real-world scenarios.  
% Fig.~\ref{fig:dataset} illustrates some image samples from our \textbf{DPED50K}. 
The vast complex scenes involved in \textbf{DPED50K} make it challenging and well suited to conduct research and evaluate the performance of image inpainting techniques with or without guidance reference images.  

\subsection{Dataset Construction}
For each input-reference image pair of size $W \times H$, we first divide them into four equal-size non-overlapping $W/2 \times H/2$ sub-images as well as crop one $W/2 \times H/2$  sub-image from their centers, resulting in five sub-images of the same size. This step aims to reduce the computation for subsequent SIFT feature matching while preserving as much as possible the important details and discriminant features of the two images in a pair. 
Next, two SIFT objects are created to calculate the key points and their feature descriptors of the target and reference sub-images. With the KNN algorithm, we match the key points and filter the matched feature descriptors to determine the eligible key points by calculating the Euclidean distance. 
Finally, the input image and the reference image are obtained by cropping the center of the eligible key points in the source and target image patches, with a size of $256 \times 256$, respectively. 
In this way, we collect a dataset named \textbf{DPED50K} of 52K paired patches, including a training set of 50K pairs and a test set of 2K pairs.

\subsection{Comparisons with Existing Datasets}
Some existing datasets, such as \textbf{CelebA}\citep{Celeb}, \textbf{Paris StreetView}\citep{paris} and \textbf{Places2}\citep{places}, have been commonly used for image inpainting tasks. \textbf{CelebA} consists of images of human faces with over 180,000 training images. One model trained on this dataset may be transferred to face editing or completion tasks, but is difficult in generalizing to other scenes.  \textbf{Paris StreetView} contains 14,900 training images and 100 test images, which are mainly focused on urban buildings from street views. \textbf{Places2} contains over 8 million images from over 365 kinds of scenes, making it particularly suitable for training image inpainting models since it enables a model to learn the distributions of various natural scenes.

As explained in Sec.~\ref{sec:introduction}, the task of reference-guided image inpainting is promising and essential, since there are conditions that infer the original appearance of a damaged image or scene with numerous historical or similar images for reference. Nevertheless, to the best of our knowledge, the existing datasets including the above-mentioned ones, all lack adequate image-to-image references, making them only suitable for training single-image inpainting models. Being the first of its kind, the newly-released \textbf{DPED50K} is expected to significantly contribute to the advancement of reference-guided image inpainting, especially the restoration of corrupted scenes to their original appearance.

\tabcolsep=0.5pt
\begin{figure*}[htb]
	\centering
\footnotesize{
		\begin{tabular}{ccccccccccccc}
				\includegraphics[width=0.075\textwidth]{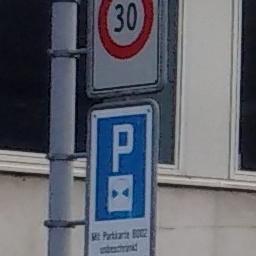} &
			\includegraphics[width=0.075\textwidth]{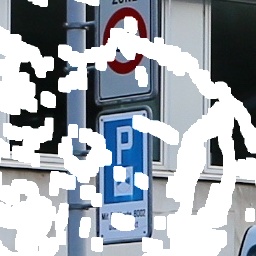} &
			\includegraphics[width=0.075\textwidth]{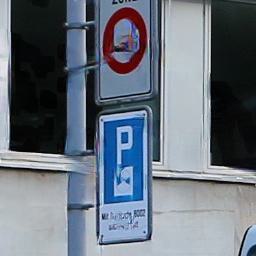} &
			\includegraphics[width=0.075\textwidth]{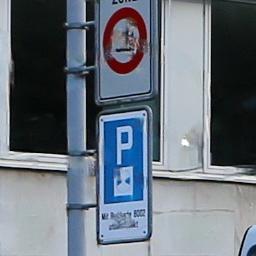} &
            \includegraphics[width=0.075\textwidth]{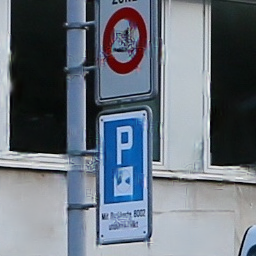} &
			\includegraphics[width=0.075\textwidth]{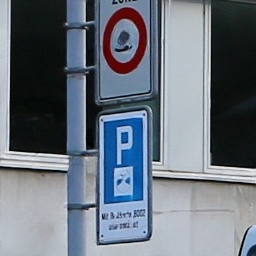} &
			\includegraphics[width=0.075\textwidth]{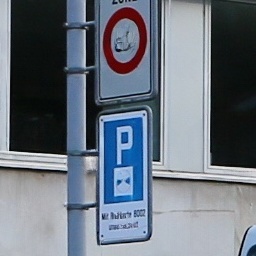} &
            \includegraphics[width=0.075\textwidth]{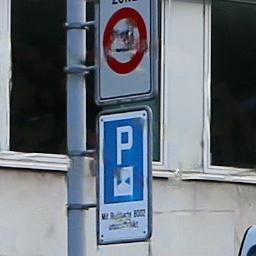} &
              \includegraphics[width=0.15\textwidth]{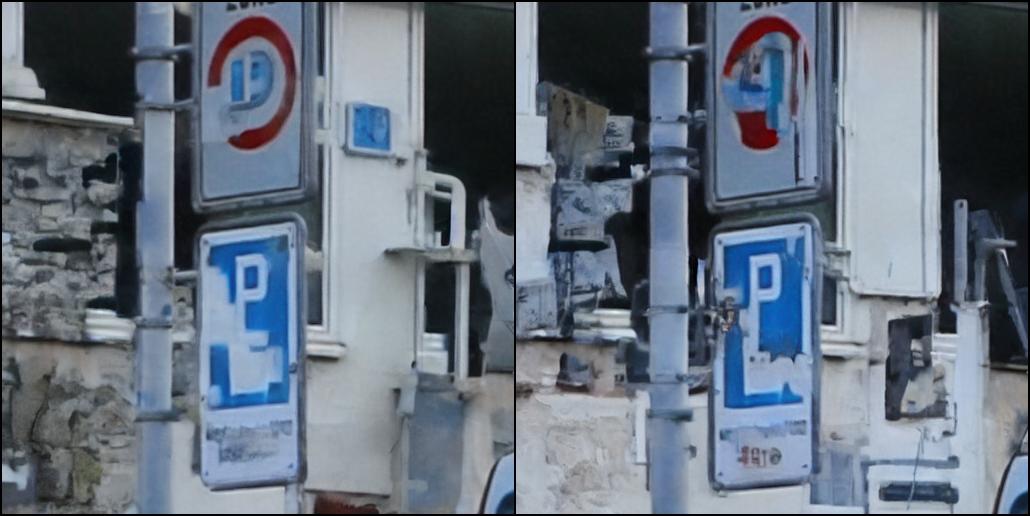} &
			\includegraphics[width=0.075\textwidth]{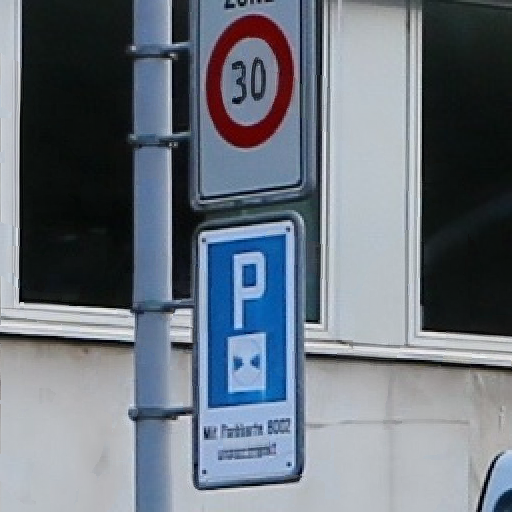} &
			\includegraphics[width=0.075\textwidth]{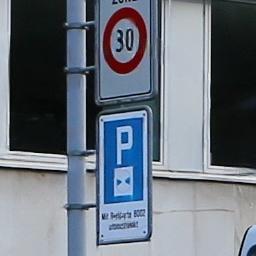} &
			\includegraphics[width=0.075\textwidth]{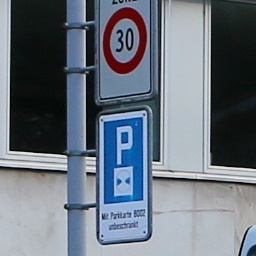} \\	
   
               \includegraphics[width=0.075\textwidth]
            {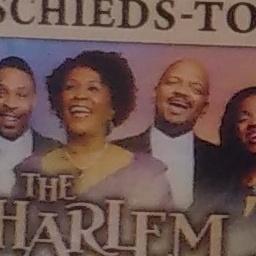} &
			\includegraphics[width=0.075\textwidth]{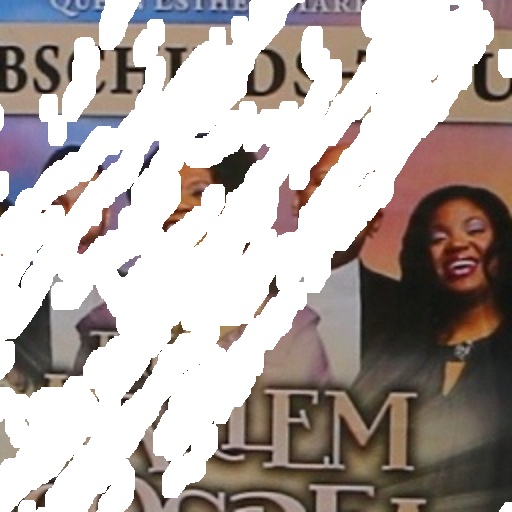} &
			\includegraphics[width=0.075\textwidth]{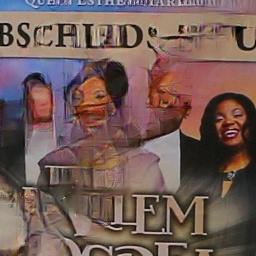} &
			\includegraphics[width=0.075\textwidth]{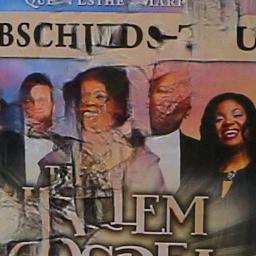} &
            \includegraphics[width=0.075\textwidth]{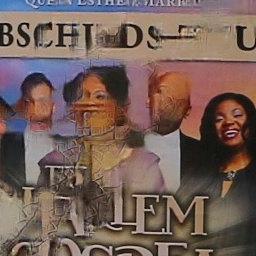} &
			\includegraphics[width=0.075\textwidth]{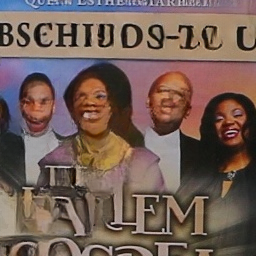} &
			\includegraphics[width=0.075\textwidth]{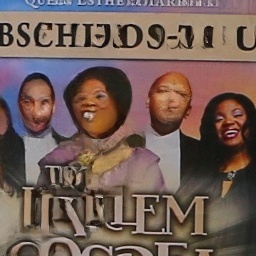} &
            \includegraphics[width=0.075\textwidth]{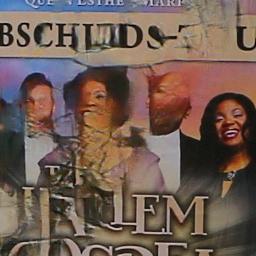} &
              \includegraphics[width=0.15\textwidth]{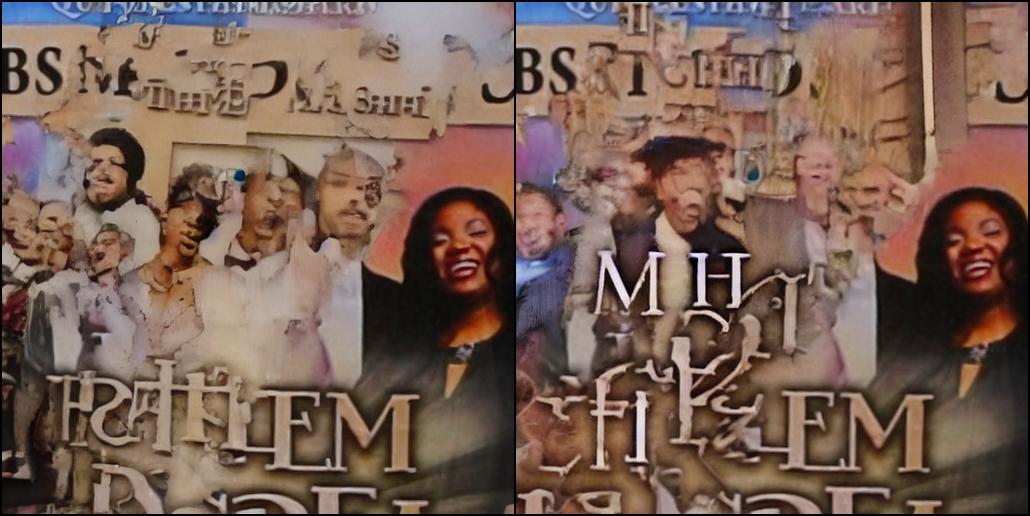} &
			\includegraphics[width=0.075\textwidth]{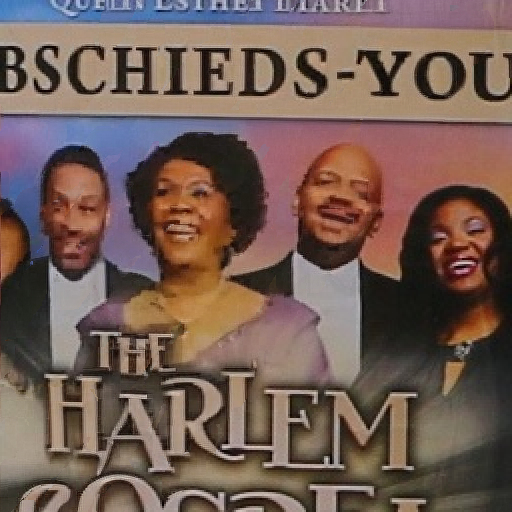} &
           \includegraphics[width=0.075\textwidth]{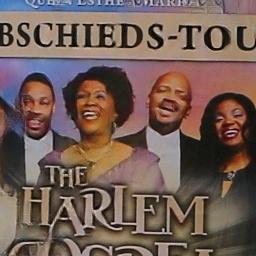} &
			\includegraphics[width=0.075\textwidth]{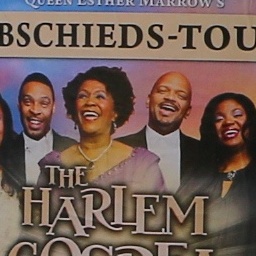} \\
			
			\includegraphics[width=0.075\textwidth]{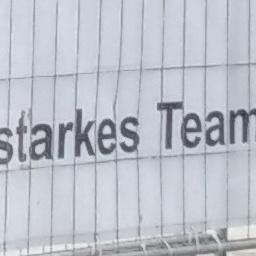} &
			\includegraphics[width=0.075\textwidth]{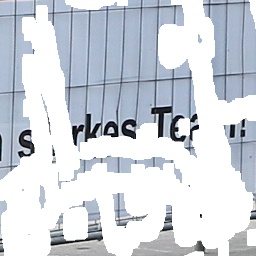} &
			\includegraphics[width=0.075\textwidth]{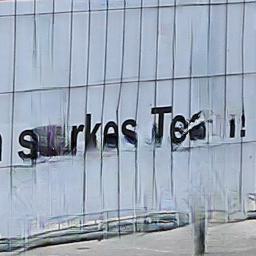} &
			\includegraphics[width=0.075\textwidth]{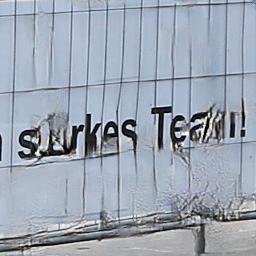} &
            \includegraphics[width=0.075\textwidth]{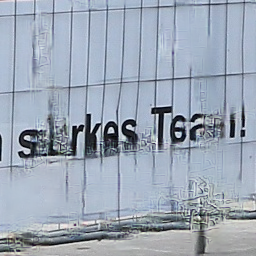} &
			\includegraphics[width=0.075\textwidth]{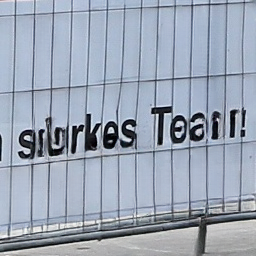} &
			\includegraphics[width=0.075\textwidth]{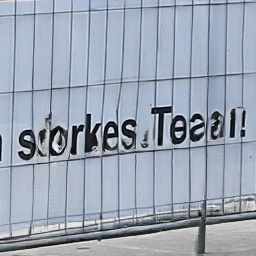} &
            \includegraphics[width=0.075\textwidth]{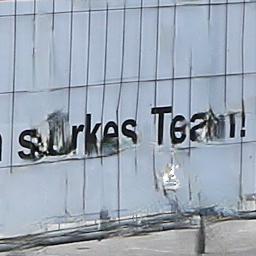} &
               \includegraphics[width=0.15\textwidth]{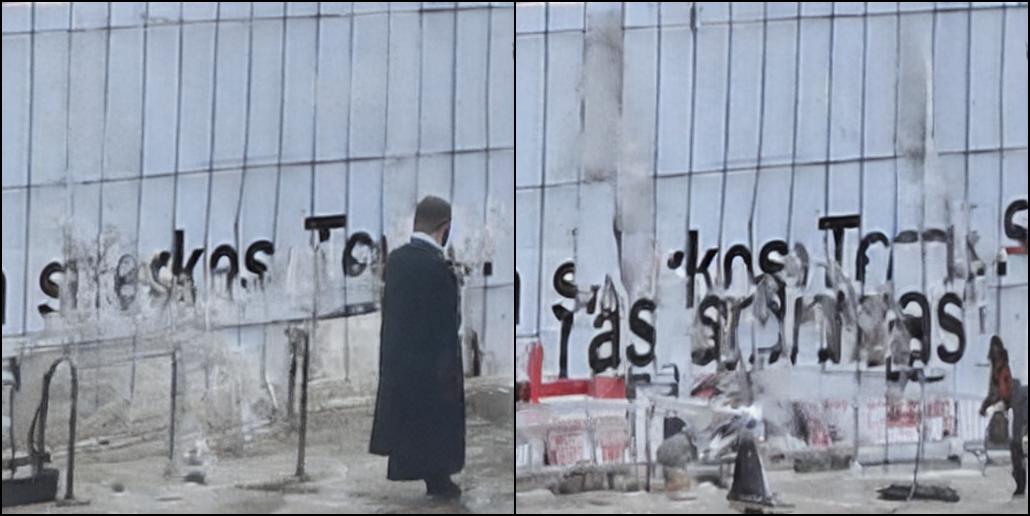} &
			\includegraphics[width=0.075\textwidth]{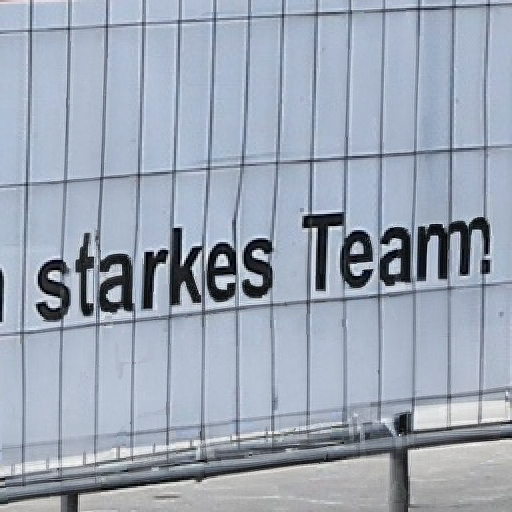} &
               \includegraphics[width=0.075\textwidth]{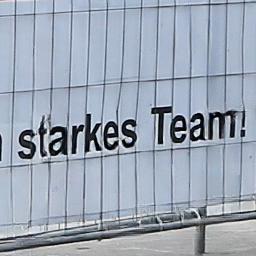} &
			\includegraphics[width=0.075\textwidth]{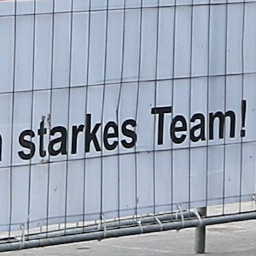} \\

                \includegraphics[width=0.075\textwidth]{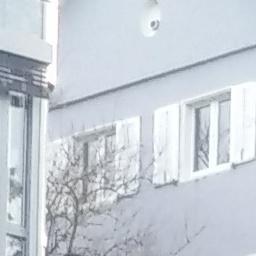} &
			\includegraphics[width=0.075\textwidth]{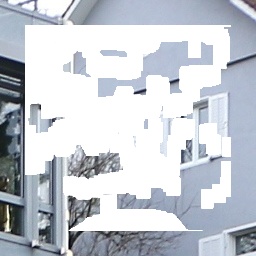} &
			\includegraphics[width=0.075\textwidth]{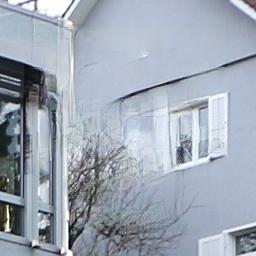} &
			\includegraphics[width=0.075\textwidth]{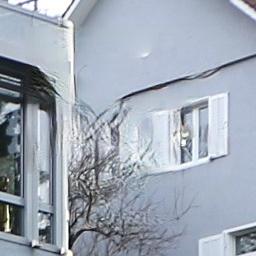} &
            \includegraphics[width=0.075\textwidth]{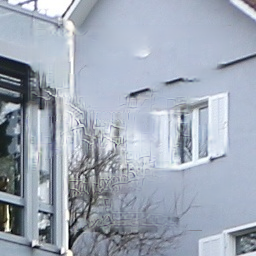} &
			\includegraphics[width=0.075\textwidth]{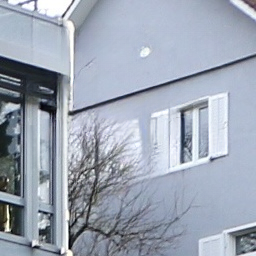} &
			\includegraphics[width=0.075\textwidth]{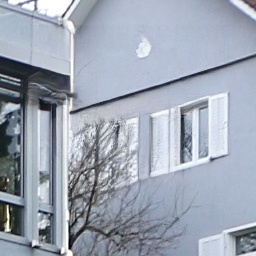} &
            \includegraphics[width=0.075\textwidth]{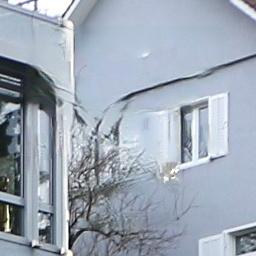} &
               \includegraphics[width=0.15\textwidth]{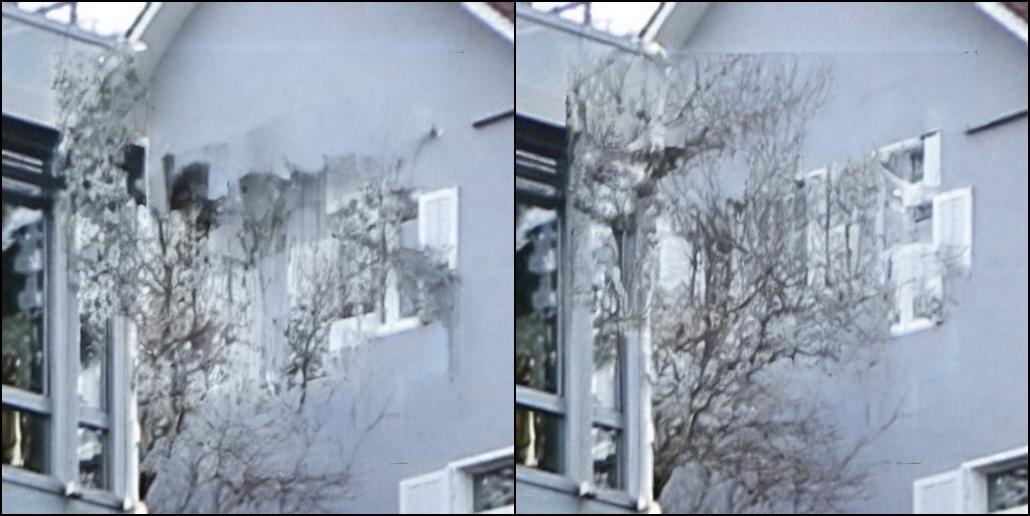} &
			\includegraphics[width=0.075\textwidth]{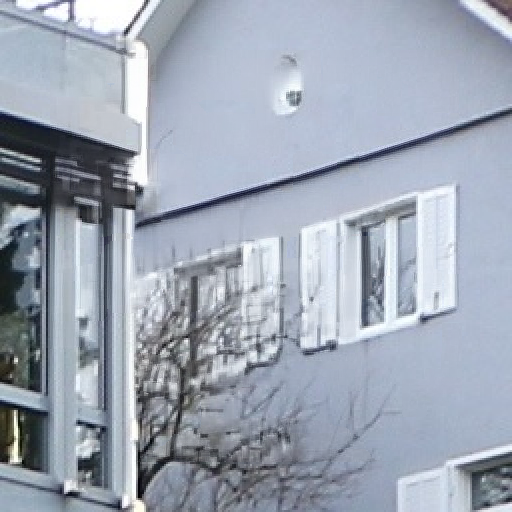} &
			\includegraphics[width=0.075\textwidth]{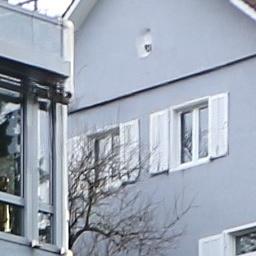} &
			\includegraphics[width=0.075\textwidth]{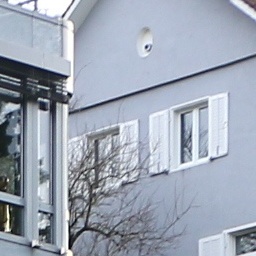} \\
   
		    \includegraphics[width=0.075\textwidth]{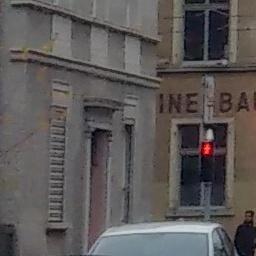} &
			\includegraphics[width=0.075\textwidth]{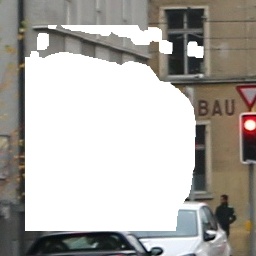} &
			\includegraphics[width=0.075\textwidth]{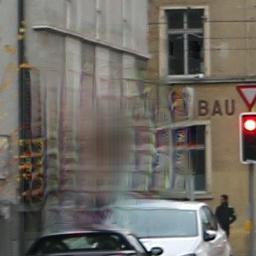} &
			\includegraphics[width=0.075\textwidth]{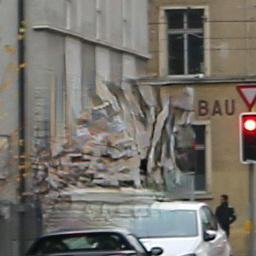} &
            \includegraphics[width=0.075\textwidth]{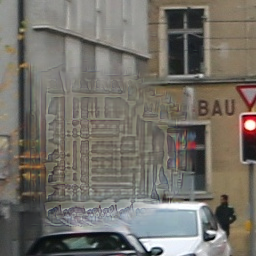} &
			\includegraphics[width=0.075\textwidth]{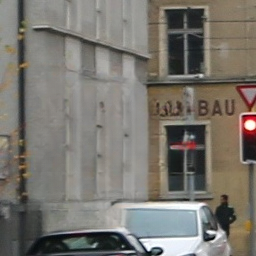} &
			\includegraphics[width=0.075\textwidth]{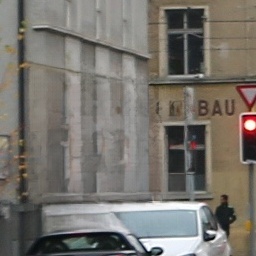} &
            \includegraphics[width=0.075\textwidth]{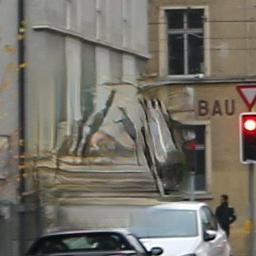} &
              \includegraphics[width=0.15\textwidth]{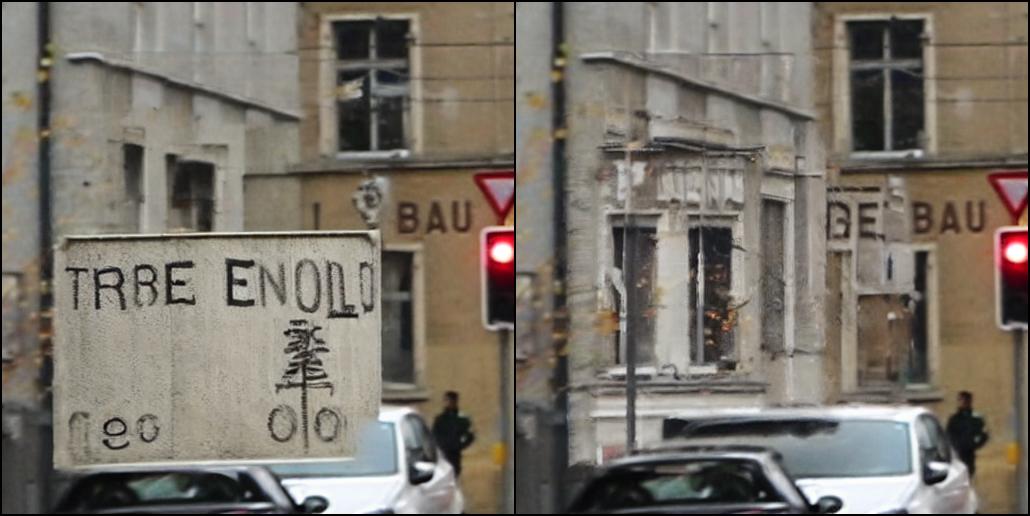} &
			\includegraphics[width=0.075\textwidth]{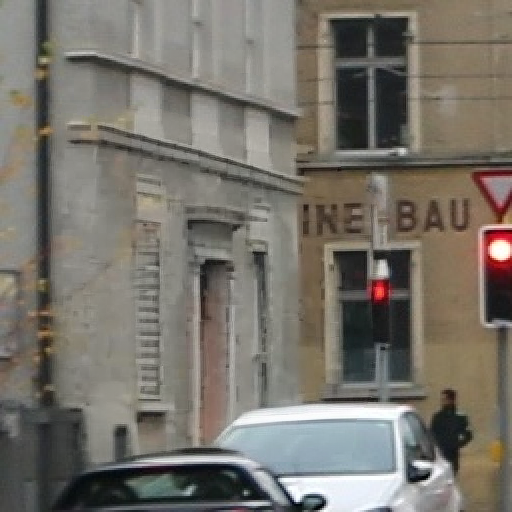} &
			\includegraphics[width=0.075\textwidth]{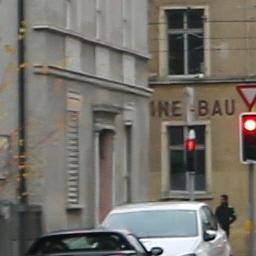} &
			\includegraphics[width=0.075\textwidth]{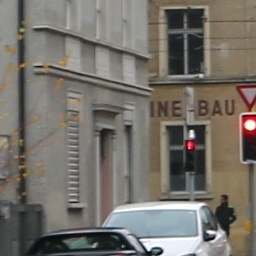} \\
   
			\includegraphics[width=0.075\textwidth]{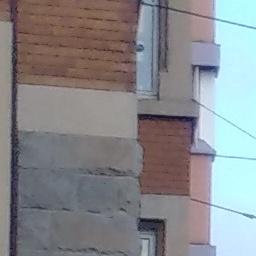} &
			\includegraphics[width=0.075\textwidth]{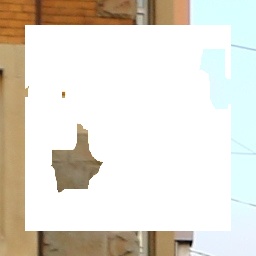} &
			\includegraphics[width=0.075\textwidth]{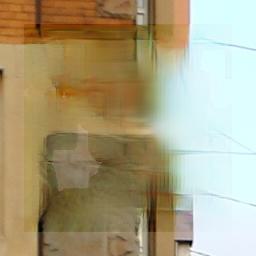} &
			\includegraphics[width=0.075\textwidth]{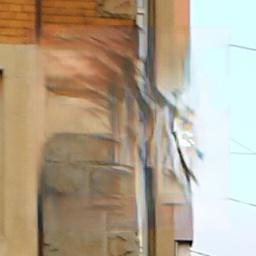} &
            \includegraphics[width=0.075\textwidth]{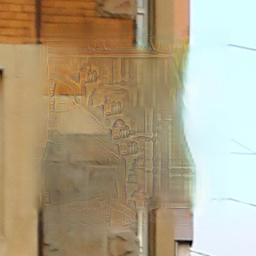} &
			\includegraphics[width=0.075\textwidth]{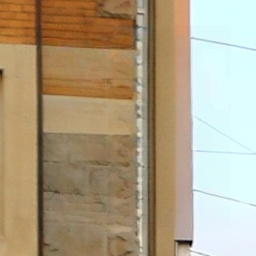} &
			\includegraphics[width=0.075\textwidth]{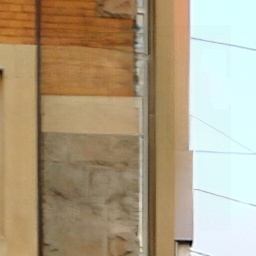} &
              \includegraphics[width=0.075\textwidth]
              {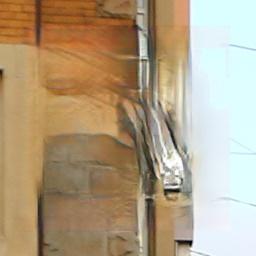} &
              \includegraphics[width=0.15\textwidth]{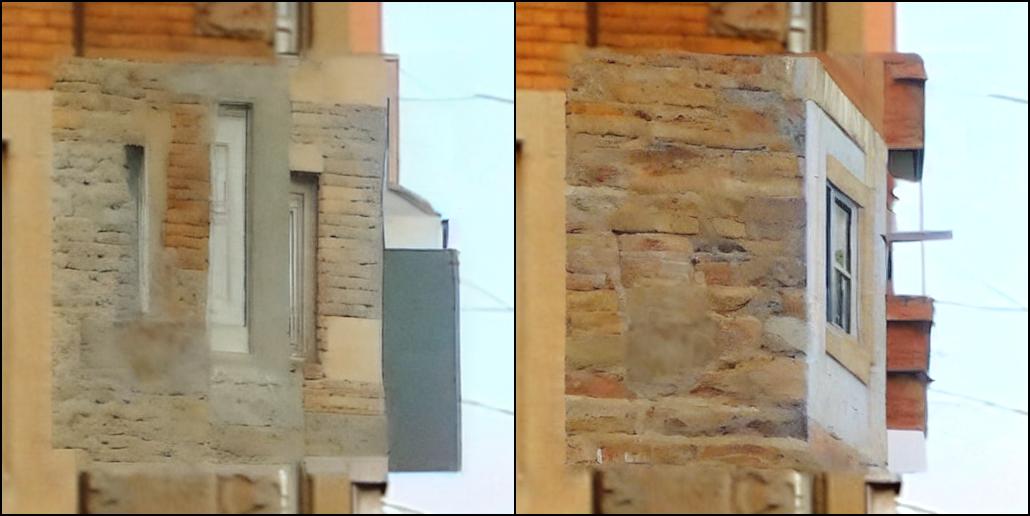} &
			\includegraphics[width=0.075\textwidth]{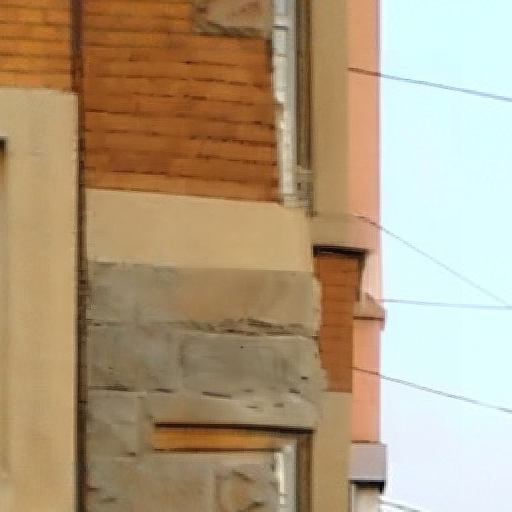} &
			\includegraphics[width=0.075\textwidth]{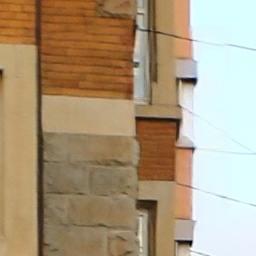} &
			\includegraphics[width=0.075\textwidth]{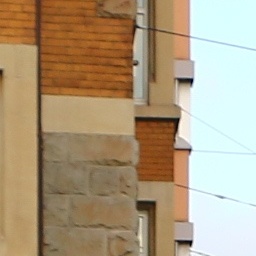} \\

			\scriptsize{(a)Ref} & \scriptsize{(b)Input} & \scriptsize{(c)EC} & \scriptsize{(d)MEDFE} & 
            \scriptsize{(e)CTSDG} & 
            \scriptsize{(f)LaMa} & \scriptsize{(g)ZITS} & 
           \scriptsize{(h)RGTSI} & 
           \scriptsize{(i)Uni-paint(exemplar)} & 
           \scriptsize{(j)LeftRefill}  &
           \scriptsize{(k)Ours} & \scriptsize{(l)GT}   \\
	\end{tabular}}
   \caption{Qualitative comparisons of inpainting results on image samples from \textbf{DPED50K}. Best viewed by zoom-in on screen.}
\label{fig:qualitive}
\end{figure*}

\section{Experimental Results}
\label{sec:experiment}
In this section, we first describe the experimental settings, including datasets, baseline methods, and implementation details. 
Next, performance comparisons between our method and baseline methods are conducted. Then, we study the performance of the proposed key modules. Finally, we apply \textbf{TransRef} to unseen real-world images to evaluate its generalization ability and discuss some failure cases.

\subsection{Experimental Settings}
\subsubsection{Implementation Details}
We implement \textbf{TransRef} using the Pytorch toolbox and optimize it using the Adam  optimizer with a learning rate of $2\times10^{-4}$ following~\citep{RGTSI}. In all experiments, we use a batch size of 32 on a single Tesla V100 GPU. The loss weights $\lambda_{l_1}$, $\lambda_{p}$, and $\lambda_{s}$ are empirically set to 1, 0.1, and 250, respectively. The training procedure for our proposed model is stopped after 400 epochs.

\subsubsection{Datasets}
We mainly evaluate \textbf{TransRef} on the proposed \textbf{DPED50K} dataset. We follow the manner suggested in  \citep{PConv} to generate irregular masks, which are separated into six categories (0--10\%, 10--20\%, ..., 50--60\%) based on the ratios of their hole sizes relative to the complete image.  Each category has 1,000 masks with damaged boundaries and 1,000 masks without damaged boundaries. All images are resized to $256\times256$ for training and testing. We also test the generalization ability of our trained models on real-world applications, \textit{i.e.}, object removal from real photos, and cloud removal from satellite images.

\subsubsection{Compared Methods}

We compare \textbf{TransRef} with inpainting methods without reference image, including EC~\citep{EC}, MEDFE~\citep{MEDFE}, CTSDG~\citep{CTSDG}, LaMa~\citep{lama}, ZITS~\citep{ZITS} and inpainting methods with reference image, including RGTSI~\citep{RGTSI}, Uni-paint ~\citep{yang2023uni} (exemplar-based), and LeftRefill~\citep{cao2024leftrefill}.
For a fair comparison, we follow the experimental settings described in the original papers for all baseline methods, except for Uni-paint~\citep{yang2023uni} and LeftRefill~\citep{cao2024leftrefill}. This exception is due to the fact that Uni-paint~\citep{yang2023uni} and LeftRefill~\citep{cao2024leftrefill} are based on the stable diffusion model~\citep{Rombach_2022_CVPR}, which inherently possesses strong generalization capabilities.

We list all the baselines with their abbreviation and a brief introduction as follows:

\textbf{EC}~\citep{EC}: Two-step inpainting with edges as low-level structural information.

\textbf{MEDFE}~\citep{MEDFE}: A mutual encoder-decoder inpainting network with feature equalizations for structure and texture completion.

\textbf{CTSDG}~\citep{CTSDG}: Two subtasks inpainting with conditional texture and structure dual generation.

\textbf{LaMa}~\citep{lama}: A resolution-robust inpainting method for large mask with fourier convolutions.

\textbf{ZITS}~\citep{ZITS}: A transformer using Zero-initialized Residual Addition (ZeroRA)~\citep{zerora} to incrementally add structural information for image inpainting with masking positional encoding.

\textbf{RGTSI}~\citep{RGTSI}: An image inpainting method with the texture and structure inference guided by the reference image.

\textbf{Uni-paint}~\citep{yang2023uni}: An unified framework for multimodal-guided image inpainting with pretrained diffusion model.

\textbf{LeftRefill}~\citep{cao2024leftrefill}: Filling right canvas based on left reference images through generalized text-to-image diffusion model.

\subsubsection{Evaluation Metrics}
We adopt several standard metrics for image inpainting tasks, including (1) Peak Signal-to-Noise Ratio (PSNR), (2) Structural Similarity Index (SSIM), (3) Fréchet Inception Distance (FID) \citep{heusel2017gans}, (4) Learned Perceptual Image Patch Similarity (LPIPS)~\citep{LPIPS}, and (5) Image-to-image alignment (I2I). The PSNR and SSIM are related to low-level pixel values, whereas the FID and LPIPS are related to high-level visual perception quality. A better image inpainting method would result in higher PSNR and SSIM, but lower FID and LPIPS. Additionally, following the experiment setting of Uni-paint ~\citep{yang2023uni}, we only compare the I2I metrics with Uni-paint ~\citep{yang2023uni} to evaluate the image-image CLIP similarity in reference-guided inpainting.

\begin{table*}[h]
\centering
\caption{Quantitative comparisons of our proposed method with state-of-the-art inpainting baselines on \textbf{DPED50K} dataset. The best two scores are indicated by \textbf{bold} and \underline{underline}, respectively ($\uparrow$: Higher is better; $\downarrow$: Lower is better).}
\setlength\tabcolsep{6pt}
\resizebox{\textwidth}{!}
{
\belowrulesep=0pt
\aboverulesep=0pt
\label{sota}
\begin{tabular}{c|c|cccccc|c}
\toprule
 Metrics & Method & 0-10\%    &  10-20\%     & 20-30\%         & 30-40\%         & 40-50\%         & 50-60\%  & Average   \\

\midrule

\multirow{7}{*}{\makecell[c]{PSNR$\uparrow$ \\ /SSIM$\uparrow$}}&  EC~\citep{EC}   &  32.64/0.962 	& 28.95/0.930 	& 26.36/0.892 	& 24.35/0.850 	& 22.67/0.802 	& 20.44/0.739       & 25.90/0.863   \\
 ~ & MEDFE~\citep{MEDFE}  & 34.66/0.972  & 31.35/0.951 	& 28.35/0.920 	& 25.92/0.881 	& 23.93/0.835 	& 21.05/0.755  & 27.54/0.886  \\
~ & CTSDG~\citep{CTSDG} &  38.35/0.987  & 32.07/0.962  & 28.56/0.927    & 26.11/0.887   & 24.22/0.844   & 21.60/0.772         & 28.48/0.897  \\
~ & LaMa~\citep{lama}  & \underline{38.97}/\underline{0.988}  & 33.11/\underline{0.966} 	& 29.83/0.937 	& 27.55/0.906 	& 25.68/0.870 	& 23.08/\underline{0.814}     &  29.70/\underline{0.914}               \\
~ & ZITS~\citep{ZITS} & \textbf{39.39}/\textbf{0.989}  &\underline{33.48}/\textbf{0.968} &30.09/\underline{0.939} 	&27.72/\underline{0.907} &25.82/\underline{0.871} 	 &23.05/0.810     & \underline{29.92}/\underline{0.914}      \\

\cline{2-9}

 ~ & RGTSI~\citep{RGTSI}  & 34.76/0.972  & 31.39/0.952 	& 28.47/0.922 	& 26.07/0.885	& 24.13/0.843 	& 21.24/0.767    &  27.68/0.890   \\
% 添加对比方法

% ~& \textcolor{blue}{Uni-paint~\citep{yang2023uni}} & / 	& /	& / 	&  	&  	 &              &       \\

~& LeftRefill~\citep{cao2024leftrefill} & 38.44/0.983 	& 33.21/0.957	& \underline{30.33}/0.927 	& \underline{28.19}/0.894 	& \underline{26.31}/0.856 	 &   \underline{23.06}/0.791          &  \underline{29.92}/0.901     \\
 
~ & \textbf{TransRef}& 35.81/0.976  &\textbf{33.83}/\underline{0.966} &\textbf{31.95}/\textbf{0.951} &\textbf{30.35}/\textbf{0.934} &\textbf{28.89}/\textbf{0.914}  &\textbf{26.90}/\textbf{0.881}    & \textbf{31.29}/\textbf{0.937}  \\

\midrule

\multirow{7}{*}{\makecell[c]{FID$\downarrow$ \\ /LPIPS$\downarrow$}}   & EC~\citep{EC}& 3.83/0.049 	& 8.41/0.080 	& 14.45/0.117 	& 22.33/0.159 	& 34.05/0.209 	& 52.46/0.288          & 22.59/0.150      \\
~& MEDFE~\citep{MEDFE}  & 2.42/0.039 	& 5.35/0.061  	& 10.44/0.093 	& 17.46/0.133 	& 27.79/0.179 	& 49.47/0.262      & 18.82/0.128        \\
~& CTSDG~\citep{CTSDG} & 1.93/0.016     & 5.95/0.047     & 12.53/0.091    & 21.96/0.143    & 34.65/0.200    & 58.93/0.295          &  22.66/0.132        \\
~& LaMa~\citep{lama} & \underline{1.18}/\underline{0.011} 	& 3.12/\underline{0.028} 	& 5.59/\underline{0.051} 	& \underline{8.21}/\underline{0.076}     & \underline{11.04}/\underline{0.105}     & \underline{16.30}/\underline{0.161}              &  \underline{7.57}/\underline{0.072} \\
~& ZITS~\citep{ZITS}& \textbf{1.08}/\textbf{0.010} 	 &\underline{2.98}/\textbf{0.027}     & \underline{5.42}/\textbf{0.050} 	& 8.27/\underline{0.076} 	& 11.47/0.107 	& 18.58/0.169                   &  7.97/0.073 \\

\cline{2-9}

~& RGTSI~\citep{RGTSI} & 2.49/0.039 	& 5.54/0.053 	& 10.73/0.097 	& 17.84/0.137 	& 27.83/0.183 	& 50.01/0.266             &  19.07/0.129      \\

~& LeftRefill~\citep{cao2024leftrefill} & 1.99/0.015 	& 4.69/0.036	& 7.60/0.060 	& 10.62/0.085 	& 14.06/0.114	 & 19.40/0.162     &  9.73/0.079     \\

~& \textbf{TransRef} & 1.59/0.033      &\textbf{2.53}/0.042     & \textbf{3.79}/\textbf{0.050}      & \textbf{4.96}/\textbf{0.066}     & \textbf{6.31}/\textbf{0.079}     & \textbf{8.17}/\textbf{0.101}              &  \textbf{4.56}/\textbf{0.062}   \\

\midrule
\multirow{2}{*}{\makecell[c]{I2I$\uparrow$}}   & Uni-paint~\citep{yang2023uni} & 0.9126 	& 0.9130	& 0.8466 	& 0.8291	& 0.7900 	 & 0.6845     &   0.8293    \\
~& \textbf{TransRef} & \textbf{0.9531} 	& \textbf{0.9361}	& \textbf{ 0.9424} 	& \textbf{0.9258} 	& \textbf{0.9365} & \textbf{0.9485}     &  \textbf{0.9404}    \\

\bottomrule
\end{tabular}

}
\end{table*}

\subsection{Performance Comparisons}
In this section, we compare our method qualitatively and quantitatively to the baselines, followed by a comparison of computational complexity.

\subsubsection{Qualitative Comparisons}
Fig.~\ref{fig:qualitive} compares the qualitative results of \textbf{TransRef} and the baseline inpainting methods. In general, our method can adequately reconstruct the content of the original scene. Specifically, the completed images of EC~\citep{EC} and MEDFE~\citep{MEDFE} exhibit severe artifacts and distortions in the image contents, whereas LaMa~\citep{lama} and ZITS~\citep{ZITS} can generate comparatively plausible structures and textures, but they still fail to restore the original appearances faithfully. 
As for diffusion-based methods, Uni-paint\citep{yang2023uni} requires model fine-tuning for each test image, and the addition of noise sampling during this process allows it to generate diverse restoration results. We randomly selected and displayed two of these results. It can be observed that even with guidance from the reference image, Uni-paint\citep{yang2023uni} struggles to capture sufficient contextual information, leading to incorrect structures and textures that differ significantly from the original image.  LeftRefill\citep{cao2024leftrefill} effectively utilizes the reference image information to restore the damaged scene more accurately. However, when dealing with large masks, LeftRefill\citep{cao2024leftrefill} still struggles to restore the walls of buildings to their original state (as seen in the sixth row), indicating that diffusion models still face difficulties in capturing the contextual information of the image fully.

In contrast, the images generated by  \textbf{TransRef} are superior in two aspects: 1) better shapes and sharper edges (\emph{e.g.}, the characters in the third row and the windows in the fourth row); and 2) novel patterns in the contents (\emph{e.g.}, the faces in the second row and the building wall in the last row, whose patterns are not similar to those of their neighboring contents). The reference images can provide sufficient realistic guidance for filling in the corrupted area.

\subsubsection{Quantitative Comparisons}
We perform quantitative evaluations with various mask ratios ranging from 0\% to 60\% on \textbf{DPED50K}. As shown in Tab.~\ref{sota}, it is evident that \textbf{TransRef} outperforms all baseline methods, and the performance gain increases as the mask ratio increases. In the case of large masks ($\geq$ 40\%), the PSNR performance gain with \textbf{TransRef} is greater than 3dB, justifying our observation from the qualitative comparisons that the reference images provide sufficient information in restoring the pixel-wise correct results. Additionally, following the experiment setting of Uni-paint ~\citep{yang2023uni}, we only compare the I2I metrics with Uni-paint ~\citep{yang2023uni} to evaluate the image-image CLIP similarity in reference-guided inpainting. Our proposed method can still excel at capturing the semantics of the reference image.

\tabcolsep=0.5pt
\begin{figure*}[t]
	\centering
\footnotesize{
		\begin{tabular}{ccccccc}
		    \includegraphics[width=0.14\textwidth]{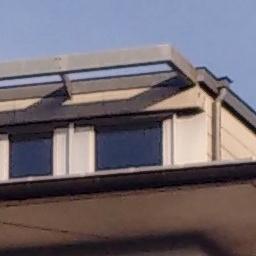} &
			\includegraphics[width=0.14\textwidth]{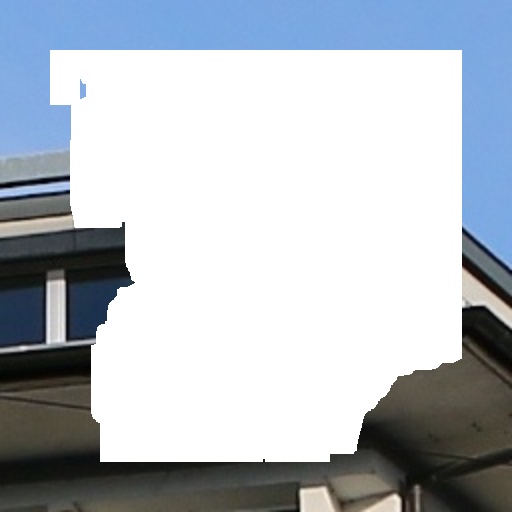} &
			\includegraphics[width=0.14\textwidth]{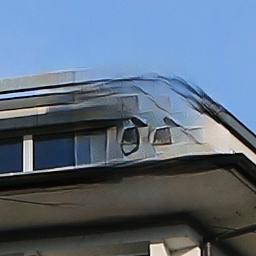} &
			\includegraphics[width=0.14\textwidth]{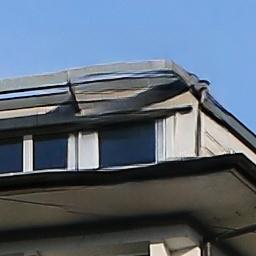} &
			\includegraphics[width=0.14\textwidth]{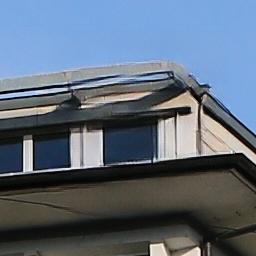} &
			\includegraphics[width=0.14\textwidth]{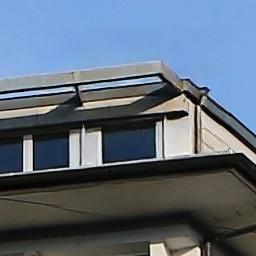} &
			\includegraphics[width=0.14\textwidth]{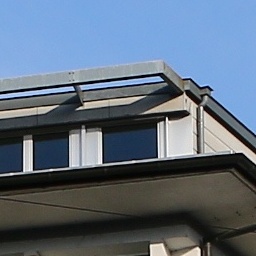}\\	
   
		    \includegraphics[width=0.14\textwidth]{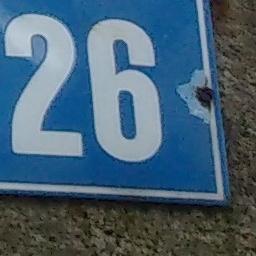} &
			\includegraphics[width=0.14\textwidth]{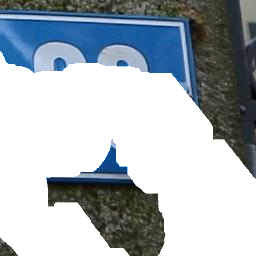} &
			\includegraphics[width=0.14\textwidth]{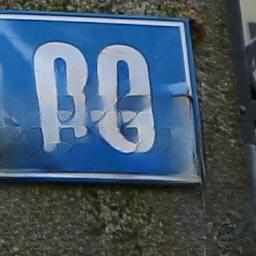} &
			\includegraphics[width=0.14\textwidth]{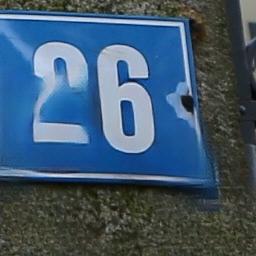} &
			\includegraphics[width=0.14\textwidth]{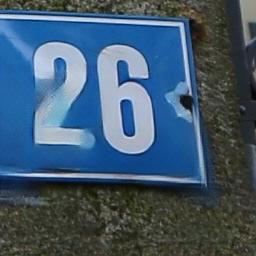} &
			\includegraphics[width=0.14\textwidth]{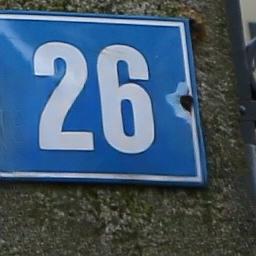} &
			\includegraphics[width=0.14\textwidth]{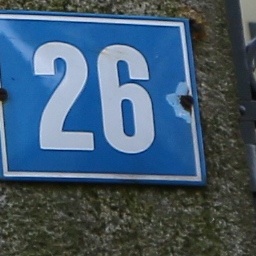}\\	
			(a) Ref & (b) Input & (c) Basic &
			(d) PA & (e) Ref-PA & (f) Ours & (g) GT \\
	\end{tabular}}
   \caption{Qualitative comparisons on four network variants to demonstrate the effects of the proposed Ref-PA and Ref-PT modules.}
	\label{network}
\end{figure*}

\tabcolsep=0.5pt
\begin{figure*}[t]
	\centering
\footnotesize{
		\begin{tabular}{ccccccccc}
		  \scriptsize{RS Score}&  0.83 & 0.74 & 1.00 & 0.45 & 0.34 & 0.00 \\
    \includegraphics[width=0.14\textwidth]{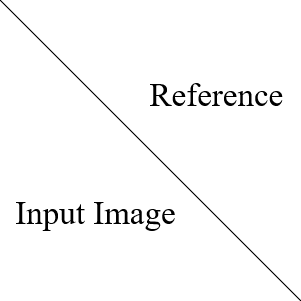} &
			\includegraphics[width=0.14\textwidth]{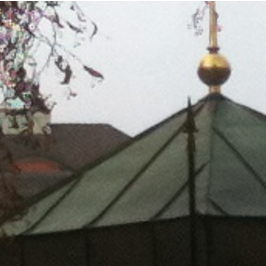} &
			\includegraphics[width=0.14\textwidth]{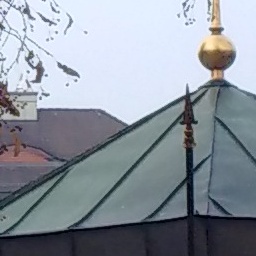} &
			\includegraphics[width=0.14\textwidth]{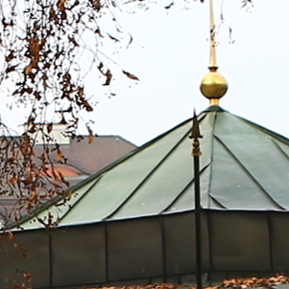} &
			\includegraphics[width=0.14\textwidth]{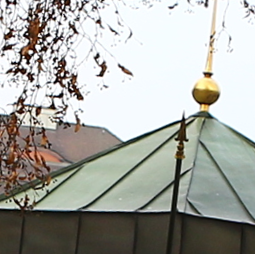}&
            \includegraphics[width=0.14\textwidth]{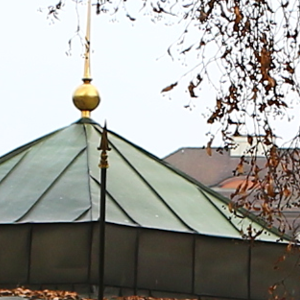}&
                \includegraphics[width=0.14\textwidth]
            {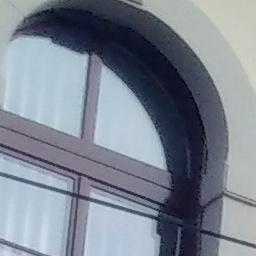}\\ 
            % I2I & 0.7993 & 0.8960 & 0.8960 & 0.8647 & 0.9194 & 0.6831 \\
			\includegraphics[width=0.14\textwidth]
		  {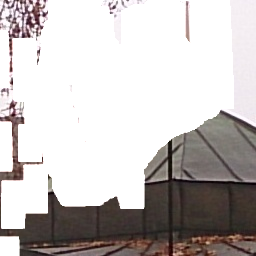} &
			\includegraphics[width=0.14\textwidth]{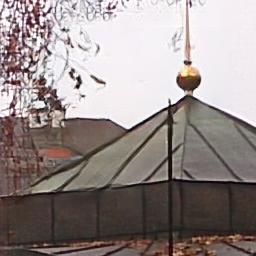} &
			\includegraphics[width=0.14\textwidth]{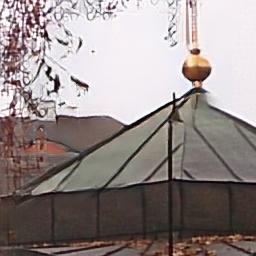} &
			\includegraphics[width=0.14\textwidth]{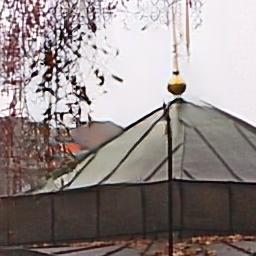} &
			\includegraphics[width=0.14\textwidth]{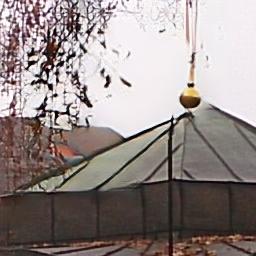} &
                \includegraphics[width=0.14\textwidth]{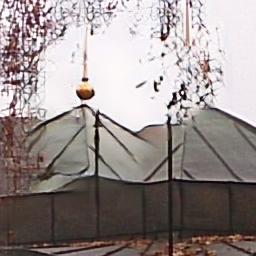} &
                \includegraphics[width=0.14\textwidth]
            {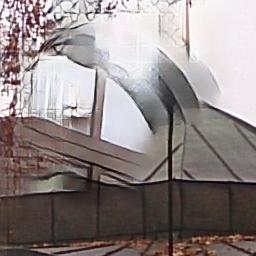}\\
			\scriptsize{(a) Input} & \scriptsize{(b) Style1} & \scriptsize{(c) Style2} & \scriptsize{(d) Style3} & \scriptsize{(e) Rotation} & \scriptsize{(f) Flipped} & \scriptsize{(g) Random} \\

	\end{tabular}}
   \caption{Qualitative comparisons on different reference variants. The first row reports the RS Score of each reference image. A higher RS Score indicates that the reference image in the batch is more similar to the input image within that batch. The second row shows reference images of different reference variants and the last row shows the corresponding completed image.}
	\label{style}
\end{figure*}

\tabcolsep=0.5pt
\begin{figure}[tb]
	\centering
		\begin{tabular}{ccccc}
			%\vspace{-0.7mm}
		\rotatebox{90}{\footnotesize{~~~~~~Scale 1}} &\includegraphics[width=0.24\columnwidth]{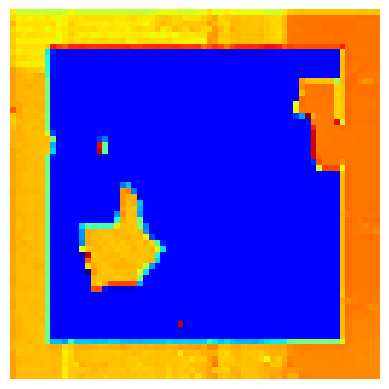} &
			\includegraphics[width=0.24\columnwidth]{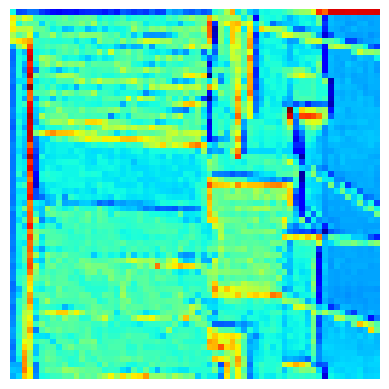} &
			\includegraphics[width=0.24\columnwidth]{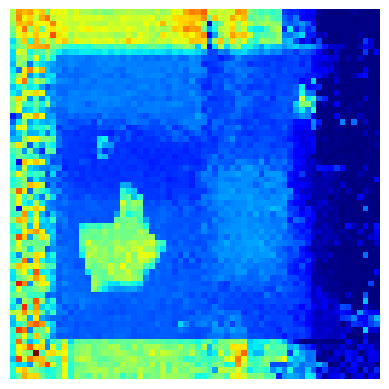} & 
                \includegraphics[width=0.24\columnwidth]{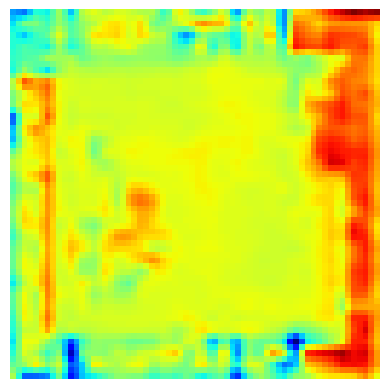}\\
	
			\rotatebox{90}{\footnotesize{~~~~~~Scale 2}} &\includegraphics[width=0.24\columnwidth]{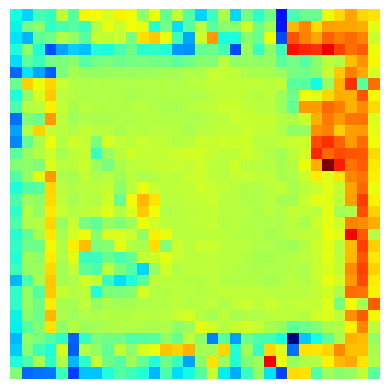} &
			\includegraphics[width=0.24\columnwidth]{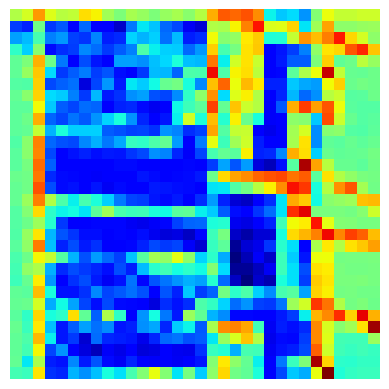} &
			\includegraphics[width=0.24\columnwidth]{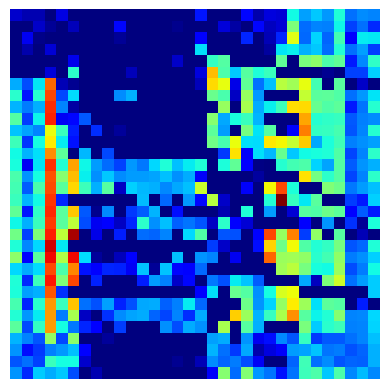} &
                \includegraphics[width=0.24\columnwidth]{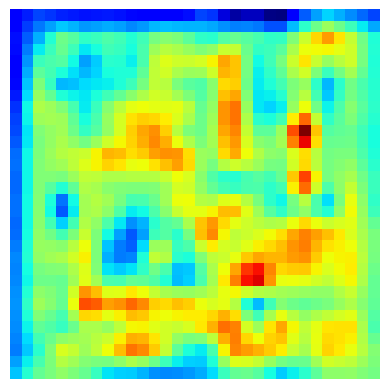}\\
		
		\rotatebox{90}{\footnotesize{~~~~~~Scale 3}} &	\includegraphics[width=0.24\columnwidth]{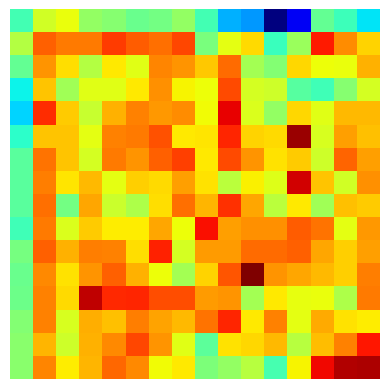} &
			\includegraphics[width=0.24\columnwidth]{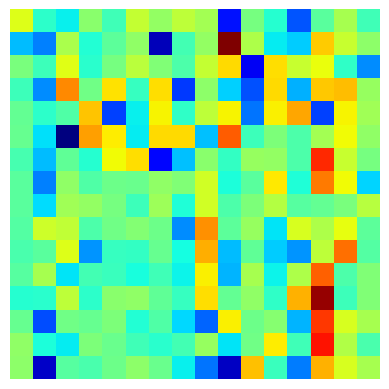} &
			\includegraphics[width=0.24\columnwidth]{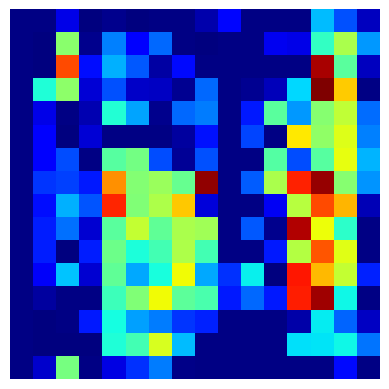} &
                \includegraphics[width=0.24\columnwidth]{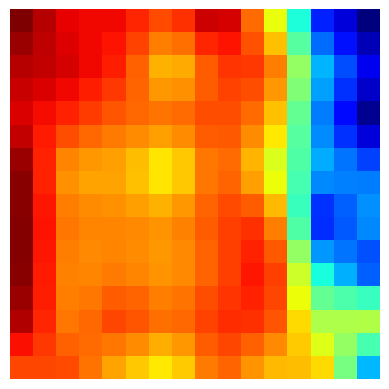}\\
		~&\small{(a)} $P_{in}$ & \small{(b) $P_{ref}$}& \small{(c) $P_{LA}$} & \small{(d) $f_{ref}$}\\
	\end{tabular}
\caption{Visualization of the feature maps of the last sample image in Fig.~\ref{fig:qualitive} derived by the reference embedding procedure in different scales. 
}
\label{vis_pam}
\end{figure}

\begin{table}[tb]
\centering
\caption{Comparison of computational complexities of different network models (\textit{Params.}:  the number of model parameters; \textit{Time}: inference or fine-tuning time, default inference time), where the two lowest complexities are highlighted by \textbf{bold} and \underline{underline}, respectively.}
\setlength\tabcolsep{6pt}

\resizebox{0.5\textwidth}{!}
{
\begin{tabular}{c|ccc}
\Xhline{1pt}
Method &  FLOPs (G)    &  Params (M)     & Time (s)       \\
\Xhline{0.6pt}
EC ~\citep{EC} 	    & 122.64	& \textbf{21.53}    & \underline{0.022} 		  \\
MEDFE ~\citep{MEDFE}  &  137.93 	& 130.32 	& 0.040 	 \\
CTSDG ~\citep{CTSDG}  & \underline{17.67} 	& 52.15	&  \textbf{0.020}  \\
LaMa ~\citep{lama}   & 98.83 	& 50.98 	& 0.056  \\
ZITS ~\citep{ZITS}   & 182.73 	& 67.90 	& 0.085  \\
\Xhline{1pt}
RGTSI ~\citep{RGTSI}   & 146.67 	& 175.55 	& 0.077  \\
Uni-paint~\citep{yang2023uni} & - 	& 859.22	& 58.234(Fine-tune)  \\
LeftRefill~\citep{cao2024leftrefill} & - 	& 1306.13	&  2.818  \\
\textbf{TransRef}    & \textbf{7.55}	    & \underline{41.97} 	& 0.026 \\
\Xhline{1pt}
\end{tabular}
}
\label{complexity}
\end{table}

\begin{table*}[tb]
\centering
\caption{Ablation study of the key modules of our proposed method on \textbf{DPED50K}, where the best two PSNR/SSIM scores are highlighted by \textbf{bold} and \underline{underline}, respectively ($\uparrow$: Higher is better; $\downarrow$: Lower is better).}
\setlength\tabcolsep{2pt}
\resizebox{1.0\textwidth}{!}
{
\belowrulesep=0pt
\aboverulesep=0pt
\begin{tabular}{cc|ccc|cccccc|c}
\toprule
 &\multirow{2}{*}{Method}&\multicolumn{2}{c}{Ref-PA} &\multirow{2}{*}{Ref-PT}&  \multirow{2}{*}{0--10\% }   &  \multirow{2}{*}{10--20\%}     & \multirow{2}{*}{20--30\%}         & \multirow{2}{*}{30--40\% }        & \multirow{2}{*}{40--50\% }        & \multirow{2}{*}{50--60\% }   &\multirow{2}{*}{Average} \\ 
 \cline{3-4} 
& &PA & PH & ~ &~&~&~&~&~&~&~\\
\midrule
(1)& Basic&\xmark & \xmark & \xmark   & 35.25/0.974  	& 32.45/0.957 	& 29.84/0.931 	& 27.62/0.900 & 25.75/0.864 	& 22.92/0.798  & 28.97/0.904  \\

(2)&PA &\cmark & \xmark & \xmark     & 35.58/\underline{0.975}    & 33.29/0.962 &	31.10/0.943 	& 29.21/0.920 	& 27.49/0.892 	& 24.87/0.837     & 30.28/0.922              \\

(3) &Ref-PA&\cmark & \cmark & \xmark      & \underline{35.63}/\textbf{0.976} 	& \underline{33.38}/\underline{0.963} 	& \underline{31.23}/\underline{0.944} 	& \underline{29.38}/\underline{0.921} 	& \underline{27.69}/\underline{0.894} 	& \underline{25.23}/\underline{0.841} & \underline{30.42}/\underline{0.923}            \\

(4)&Ours&\cmark & \cmark & \cmark      & \textbf{35.81}/\textbf{0.976}  	& \textbf{33.83}/\textbf{0.966} 	& \textbf{31.95}/\textbf{0.951}	& \textbf{30.35}/\textbf{0.934} 	& \textbf{28.89}/\textbf{0.914}& \textbf{26.90}/\textbf{0.881}   & \textbf{31.29}/\textbf{0.937}           \\
\bottomrule
\end{tabular}
}
\label{table2}
\end{table*}

\subsubsection{Complexity Analysis}
We also compare the computational complexity of \textbf{TransRef} with the baseline methods in terms of FLOPs, the number of parameters, and the inference or fine-tuning time. We measure the time on one single NVIDIA Tesla V100. As shown in Table~\ref{complexity}, \textbf{TransRef} consumes the lowest FLOPs, a modest number of parameters, and a considerably reduced inference time compared to the other baselines. Notably, since \textbf{TransRef} utilizes a transformer-based encoder-decoder network and the way for calculating multi-head attention is accelerated similar in~\citep{segformer}, the FLOPs complexity of \textbf{TransRef} are much fewer than the compared methods. Meanwhile, \textbf{TransRef} is a single-stage network, therefore its parameters are fewer than those of ZITS~\citep{ZITS}, which uses the same transformer architecture but involves a three-stage inpainting process.
Due to the fact that Uni-paint\citep{yang2023uni} and LeftRefill\citep{cao2024leftrefill} are based on diffusion models, their FLOPs calculation is influenced by the sampling steps. Consequently, we do not compute or compare their FLOPs.  Although the diffusion-based methods can generate relatively good results, their large number of parameters and long inference or fine-tuning time make them inefficient for practical deployment.

\subsection{Ablation Study}
\subsubsection{Individual Contributions of Key Modules} 
We illustrate the individual contributions of the key modules of the reference embedding procedure by removing them from  \textbf{TransRef} individually and measuring the corresponding performance drops. The settings of the variants are incrementally added as follows: (1) Basic: only the transformer architecture with the Main-PT module; (2) PA (Ref-PA $w/o$ PH): the basic model with the Ref-PA module composed of only PA block but not PH block; (3) Ref-PA: the basic model with the Ref-PA module composing of both PA block and PH block; (4) Ours (TransRef): the full model with both Ref-PA and Ref-PT modules. 

Fig.~\ref{network} illustrates that the basic model can hardly complete the big holes with correct contents, whereas the patch alignment performed in the Ref-PA module successfully fills more realistic content into the hole. Furthermore, the PH block in the Ref-PA module and the Ref-PT module respectively help  better fuse the information from the reference with the input image and correct small flaws in the details. Table~\ref{table2} demonstrates that the quantitative performance increases with the addition of proposed modules, which is consistent with the visual results.

\tabcolsep=0.5pt
\begin{figure}[tb]
	\centering
		\begin{tabular}{cccc}
			\includegraphics[width=0.245\columnwidth]{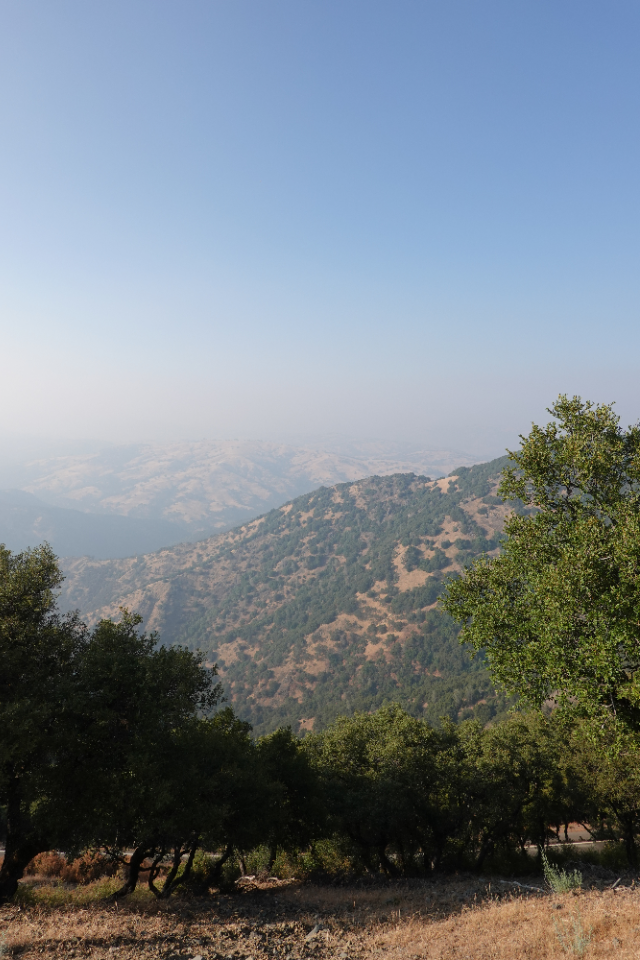} &
			\includegraphics[width=0.245\columnwidth]{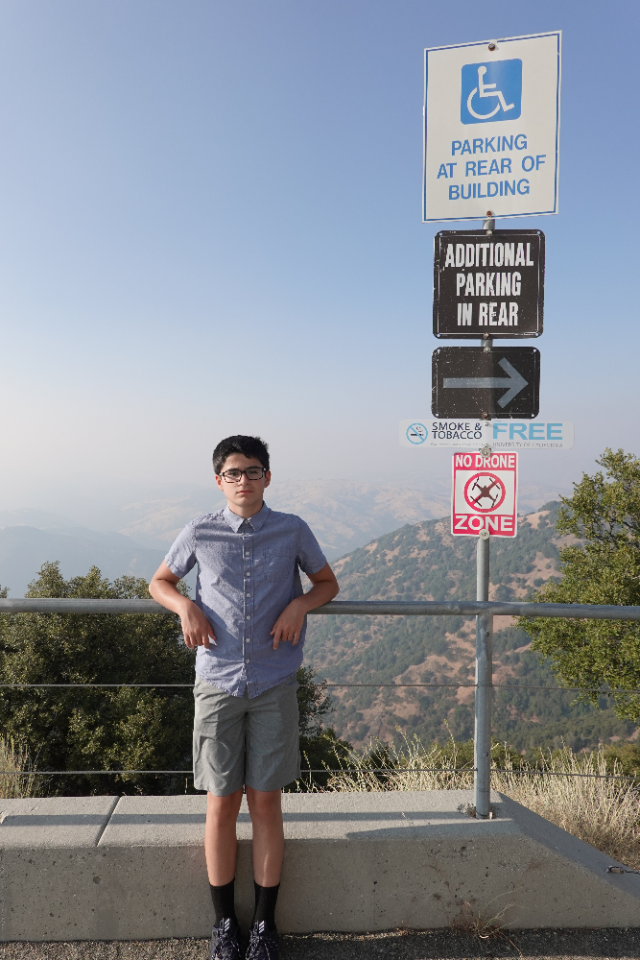} &
			\includegraphics[width=0.245\columnwidth]{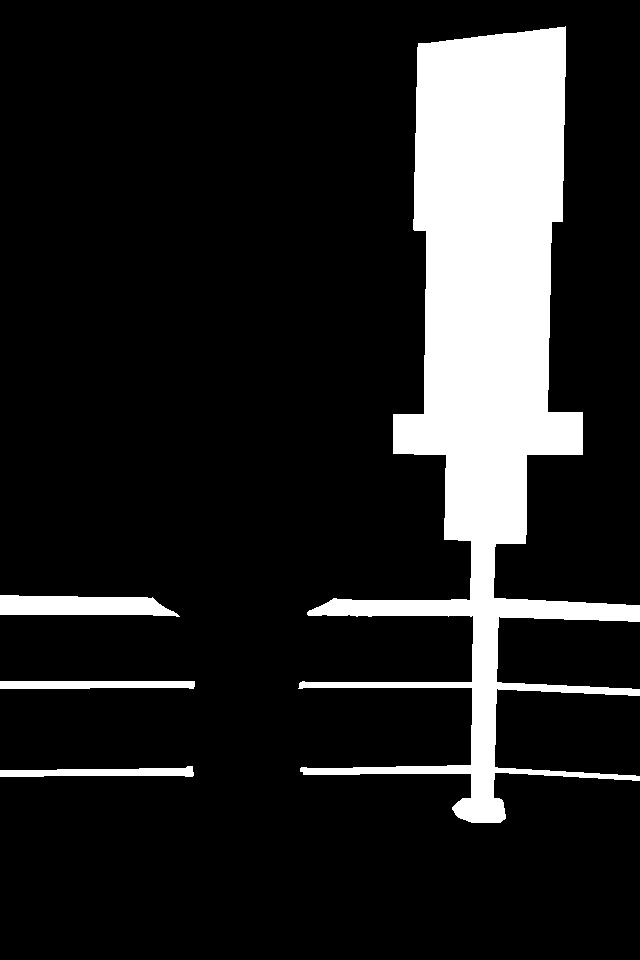} &
			\includegraphics[width=0.245\columnwidth]{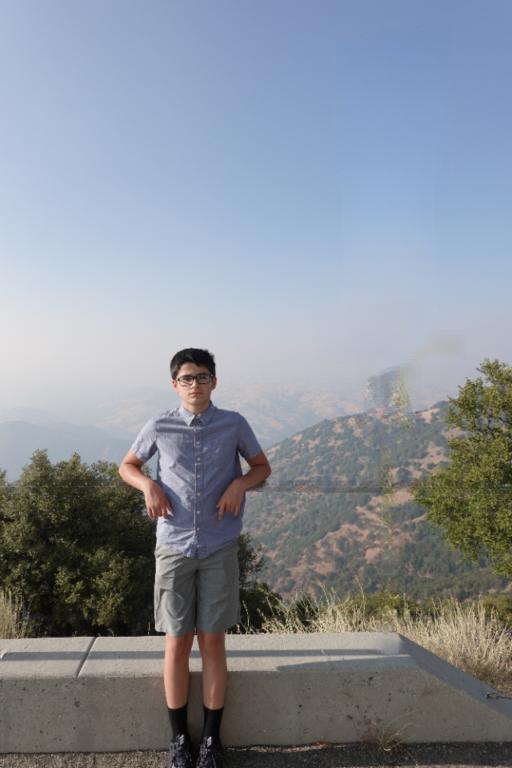} \\
   
			\includegraphics[width=0.245\columnwidth]{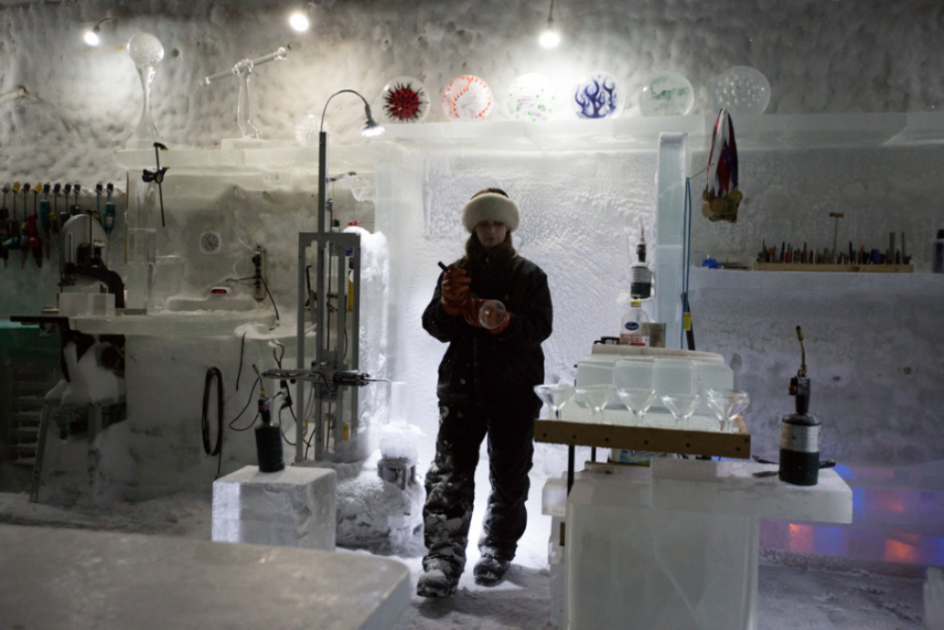} &
			\includegraphics[width=0.245\columnwidth]{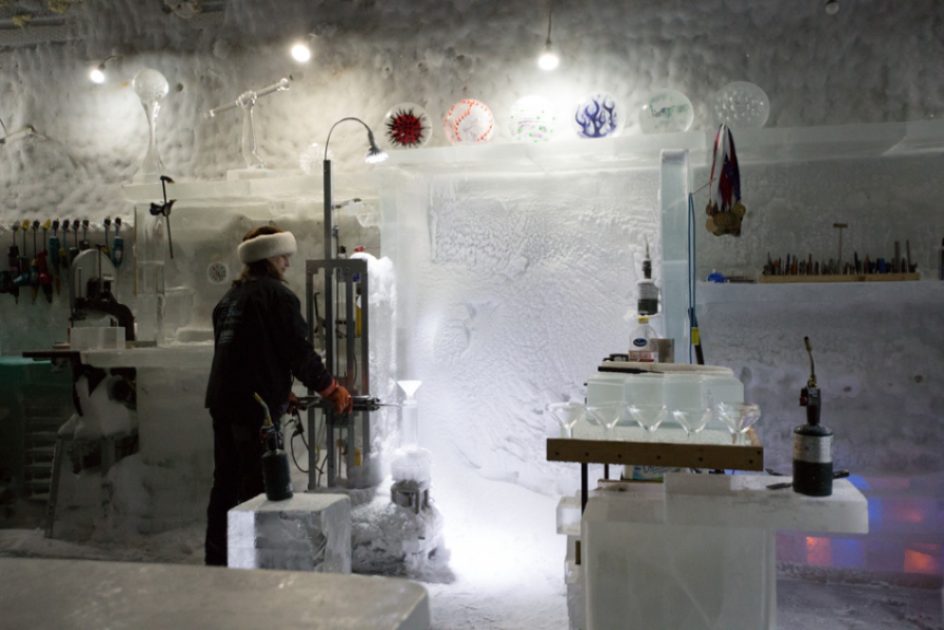} &
			\includegraphics[width=0.245\columnwidth]{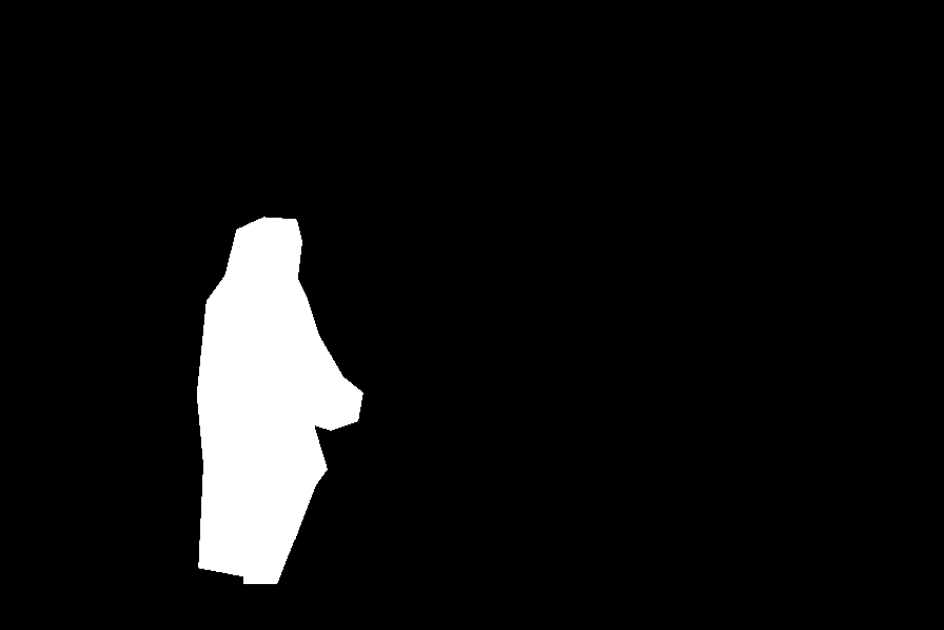} &
			\includegraphics[width=0.245\columnwidth]{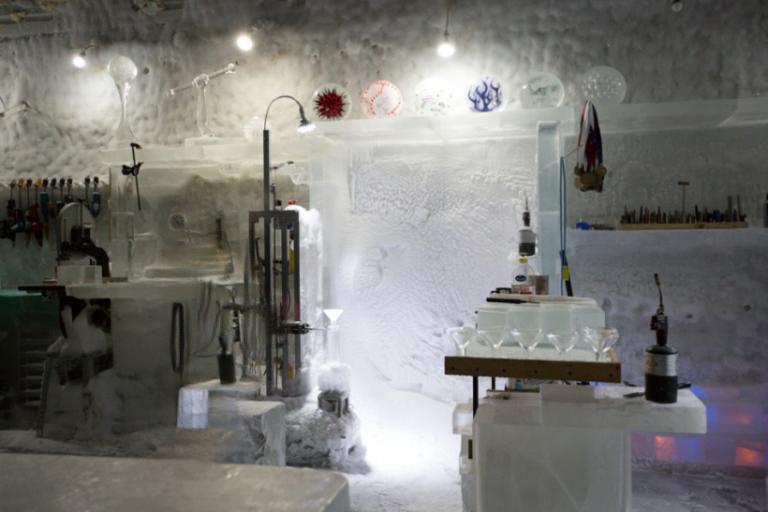}  \\
   
			\includegraphics[width=0.245\columnwidth]{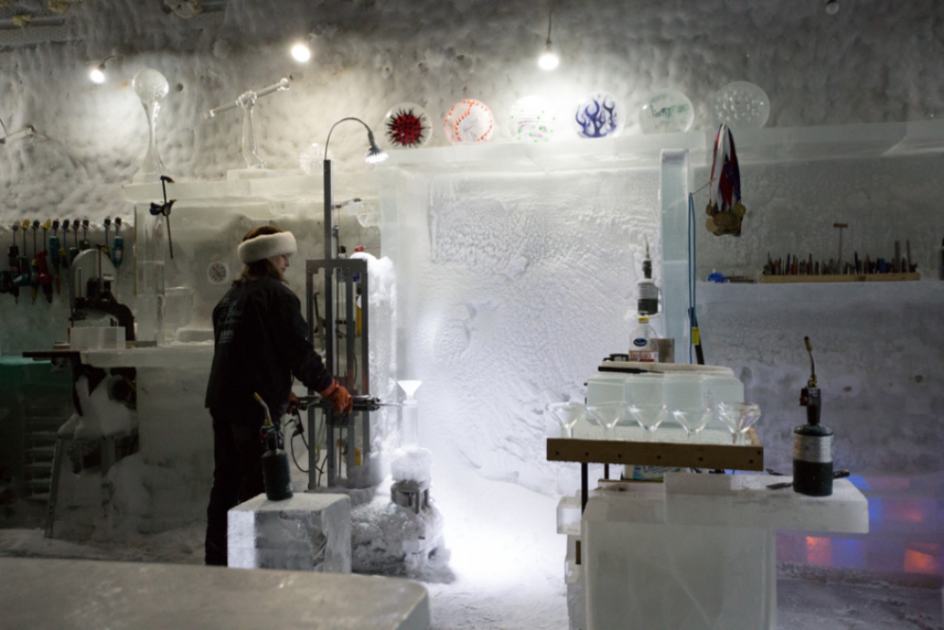} &
			\includegraphics[width=0.245\columnwidth]{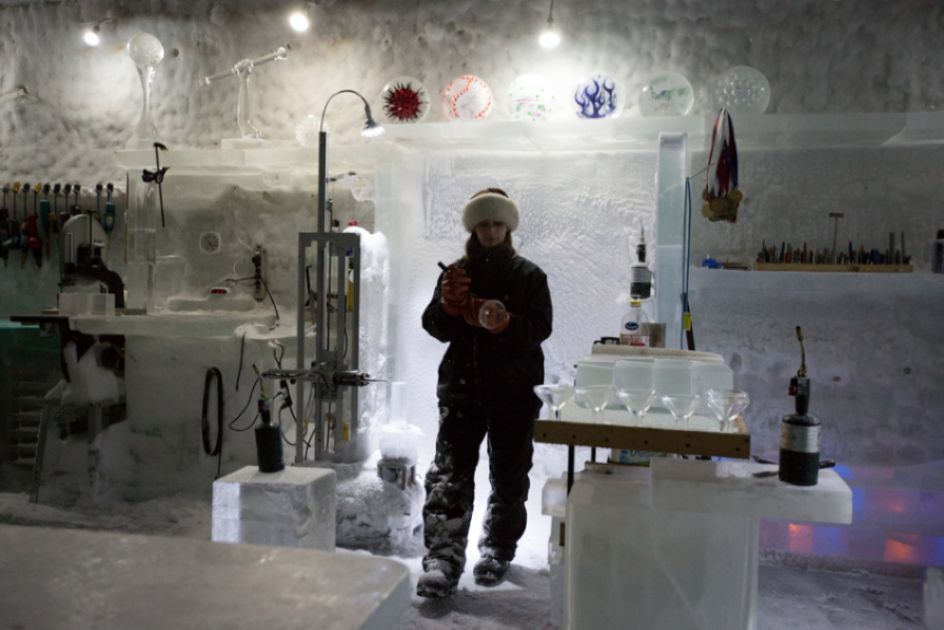} &
			\includegraphics[width=0.245\columnwidth]{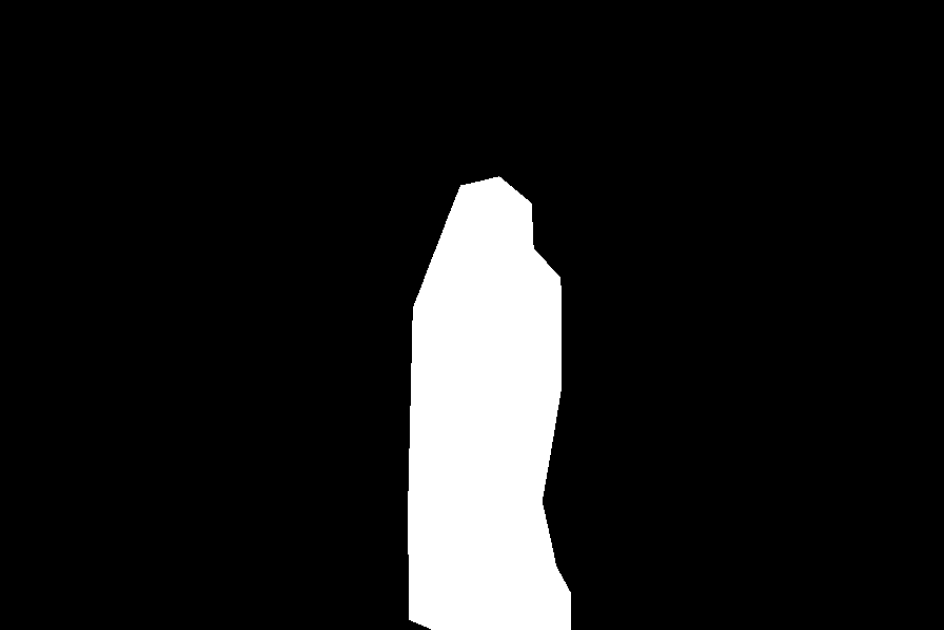} &
			\includegraphics[width=0.245\columnwidth]{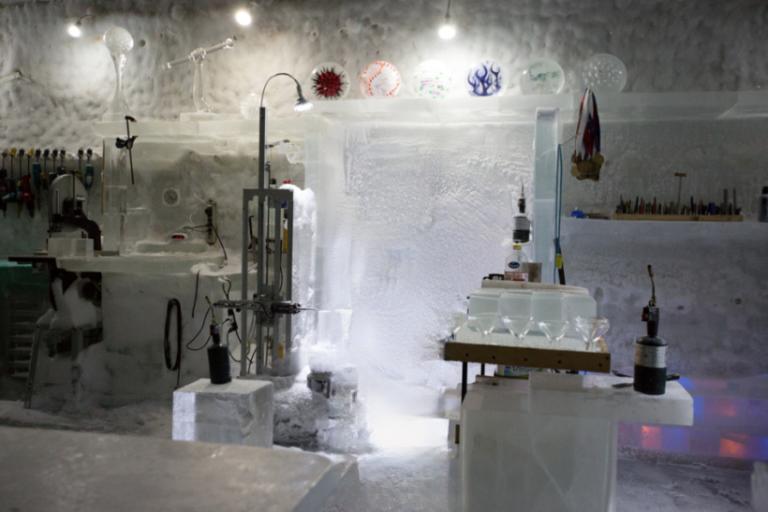} \\
    	    \footnotesize{(a) Reference} & \footnotesize{(b) Input} & 
                \footnotesize{(c) Mask} & \footnotesize{(d) \textbf{TransRef} }
	\end{tabular}
\caption{Results on the object removal from real photos using the trained \textbf{TransRef} model from \textbf{DPED50K}.}
		\label{real}
\end{figure}

\begin{figure}[tb]
	\centering
		\begin{tabular}{cccc}
            \includegraphics[width=0.245\columnwidth]{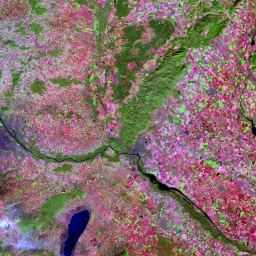} &
			\includegraphics[width=0.245\columnwidth]{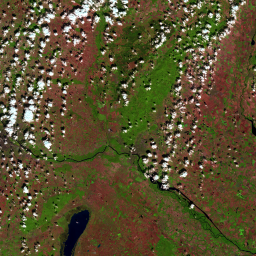}&
		  \includegraphics[width=0.245\columnwidth]{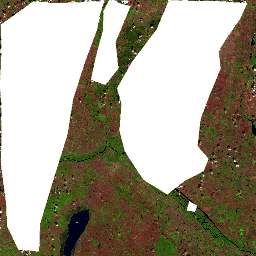} &
			\includegraphics[width=0.245\columnwidth]{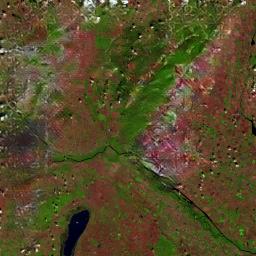} \\
   
			\includegraphics[width=0.245\columnwidth]{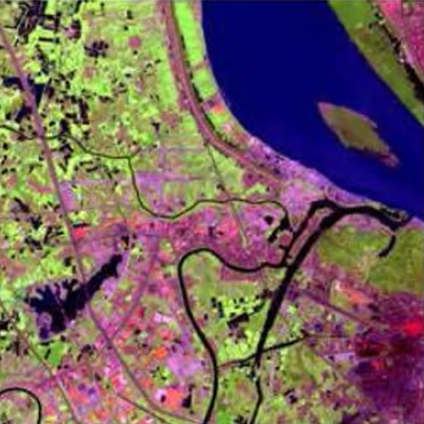} &
			\includegraphics[width=0.245\columnwidth]{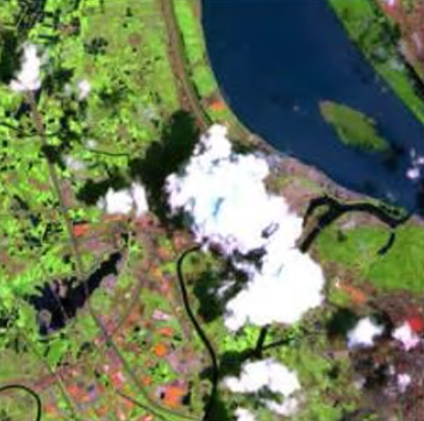} &
			\includegraphics[width=0.245\columnwidth]{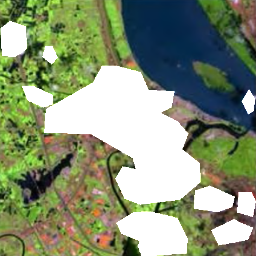} &
			\includegraphics[width=0.245\columnwidth]{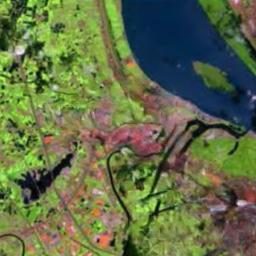}  \\
		\footnotesize{(a) Reference} & \footnotesize{(b) Cloud} & \footnotesize{(c) Input} & \footnotesize{(d) \textbf{TransRef}}
	\end{tabular}
\caption{Results on the cloud removal from satellite images using the trained \textbf{TransRef} model from \textbf{DPED50K}.}
	\label{RS}
\end{figure}

\tabcolsep=0.5pt
\begin{figure}[tb]
	\centering
		\begin{tabular}{cccc}
			\includegraphics[width=0.245\columnwidth]{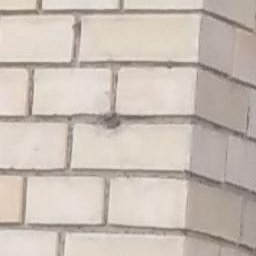} &
			\includegraphics[width=0.245\columnwidth]{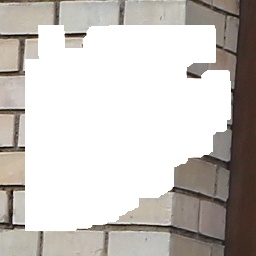} &
			\includegraphics[width=0.245\columnwidth]{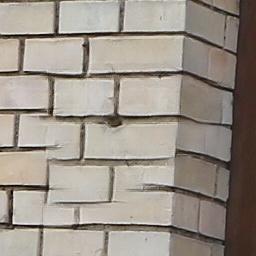} &
			\includegraphics[width=0.245\columnwidth]{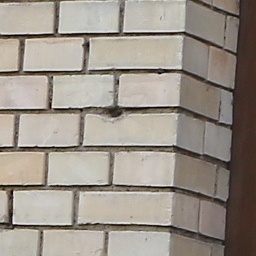} \\
                \includegraphics[width=0.245\columnwidth]{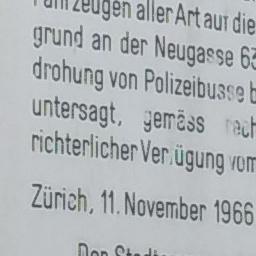} &
			\includegraphics[width=0.245\columnwidth]{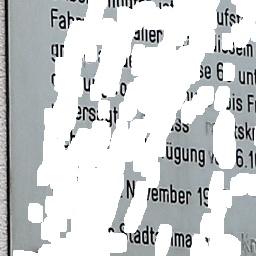} &
			\includegraphics[width=0.245\columnwidth]{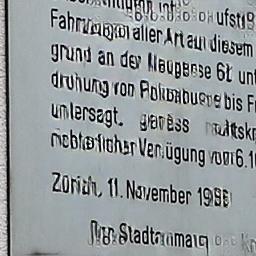} &
			\includegraphics[width=0.245\columnwidth]{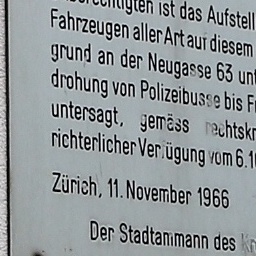} \\
       \footnotesize{(a) Reference} & \footnotesize{(b) Input} & \footnotesize{(c) \textbf{TransRef}} & \footnotesize{(d) Ground-truth}
	\end{tabular}
\caption{Examples of some failure cases with \textbf{TransRef}.}
	\label{failure}
\end{figure}

\subsubsection{Impacts of Reference Variations} 
We further study how reference variations impact the ability of \textbf{TransRef} to extract complementary information from the reference image for hole filling. The used reference variations involve three different styles of reference images, a rotated and a horizontally flipped image, and a random reference image. 
To evaluate the similarity between the reference image and the input image, based on existing evaluation metrics, we calculate the Relative Similarity Score (RS Score) by integrating PSNR and SSIM to assess which reference image in this batch is more similar to the input image within that batch. For a batch of \( N \) reference images \( I_{ref}^1, I_{ref}^2, \ldots, I_{ref}^N \) and an input image \( I_{in} \), we firstly calculate the PSNR and SSIM values and then normalize them. The RS Score combines these metrics, weighted by coefficients \( \alpha \) and \( \beta \), to balance their contributions:
\begin{align*}
\text{RS Score}_{\text{ref}}^i &= \alpha \cdot \text{Norm}\left(\text{PSNR}(I_{\text{in}}, I_{\text{ref}}^i)\right) \\
&\quad + \beta \cdot \text{Norm}\left(\text{SSIM}(I_{\text{in}}, I_{\text{ref}}^i)\right).
\end{align*}
% \[ \text{RS Score}_{ref}^i =  \alpha \cdot \text{Norm}\left(\text{PSNR}(I_{in}, I_{ref}^i)\right) + \beta \cdot \text{Norm} \left(\text{SSIM}(I_{in}, I_{ref}^i) \right). \]
A higher RS Score indicates that the reference image in the batch is more similar to the input image within that batch.

We introduced the RS Score for all reference images in this reference image batch. 
Reference images depicting correct perspectives with different styles exhibit relatively high RS Scores, whereas reference images that have undergone deformation show significantly lower scores.
As illustrated in Fig.~\ref{style}, our method can learn the crucial complementary information from the reference image in Figs.~\ref{style}(b) to (e), which all have similar textures and structures to the input image, showing that \textbf{TransRef} can fairly well tackle the reference style differences and a modest degree of rotation. 
Fig.\ref{style}(d) to (f) demonstrate that as deformation increases within images of the same style, the RS Score decreases, leading to poorer inpainting results. 
Fig.\ref{style}(g) illustrates that when the RS Score is 0, indicating no similarity between the reference image and the input image, the restoration results are at their worst. 
This highlights the importance of using reference images that closely resemble the original scene to achieve higher fidelity inpainting results.

% As in the cases of Fig.~\ref{style}(f) and (g), when the missing region cannot be well aligned with the reference, \emph{e.g.}, in the case of flipped and random references, incorrect content could be introduced from the wrong reference. 
However, it should be noted that the reference images used in the proposed method are pre-searched using SIFT features, which can eliminate the misalignment issue to some extent.

\subsubsection{Visualization of Reference Embeddings}
To demonstrate the effects of the reference embedding procedure in \textbf{TransRef}, we visualize the feature maps of a sample image before and after the reference embedding in all the scale branches in Fig.~\ref{vis_pam}. As shown in $P_{in}$ in Scale 1, the mask region is clearly shown in pure blue with no information. After the Ref-PA module, the reference feature is introduced in $P_{LA}$ and better refined in $f_{ref}$. The later scale takes the integrated feature of $f_{ref}$ and $f_{main}$ from the previous scale as the input feature and conducts the reference embedding in a coarser resolution. It is noticeable that the reference features from different scales represent different levels of information, which helps to fill in the hole gradually, and the feature in the masked region becomes more consistent with the surrounding regions. The completed features from all scales are then used for the completed image.   

\subsection{Extensive Applications}
Image inpainting is typically used in image editing tools to remove unwanted objects. Reference-guided image inpainting, on the other hand, can leverage  reference images of similar scenes to remove objects and replace them with the original scene. To evaluate the performance of our method for object removal, we apply \textbf{TransRef}  trained on \textbf{DPED50K} to the image editing of natural and remote sensing images to test its generalization to other domains.

\subsubsection{Object Removal from Real Photos} 
We test our \textbf{TransRef} model on the RealSet dataset~\citep{tranfill}, which consists of user-provided image pairs to remove objects. As shown in Fig.~\ref{real},  our model can well remove the objects and restore the original content and texture details of the background scenes.

\subsubsection{Cloud Removal from Satellite Images}
Generally, remote sensing images of landscapes are captured by satellites at different time instances. Therefore, when the landscapes in a certain place are covered by clouds, we can use same-location images from the past or future as a reference to remove the clouds. We test our model to remove the clouds from satellite images in~\citep{2020Thick} which were taken for the same landscapes at different times. As shown in Fig.~\ref{RS}, our method can effectively use multi-temporal satellite images to remove the thin clouds and thick clouds. 

\subsection{Discussion on Failure Cases}

Although our method can effectively use a reference image to fill in holes to restore their original state, it still has some limitations under some conditions, such as repeatable textures or fine structures that are difficult to align. As shown in Fig.~\ref{failure}, our model causes misalignments when the walls have dense brick joints or additional textual details, resulting in some artifacts and semantic flaws in the completed image. To address this problem, further studies on more robust feature alignment of the input and reference images are desirable. To better assess the fidelity of the completed images,  we will further look into the recent no reference quality assessment methods~\citep{Xu_2024_CVPR,chentip,wu2023q}.

\section{Conclusions}
\label{sec:conclusion}
In this paper, we explored reference-guided image inpainting, which utilizes the knowledge of a reference image to complement insufficient priors in completing complex scenes to restore realistic and original scenes when facing large masks. 
To this end, we proposed \textbf{TransRef}, an end-to-end transformer-based encoder-decoder network structure with multi-scale reference embedding, to integrate the reference information into the corrupted holes. We proposed in \textbf{TransRef} two key modules, Ref-PA and Ref-PT, to handle the problem of similar content in input and reference images that differ in geometry, style, and position. We have further constructed a new benchmark, \textbf{DPED50K}, to facilitate the development and evaluation of reference-guided inpainting methods. Both quantitative and qualitative experiments conducted on the proposed dataset demonstrate the superiority of \textbf{TransRef}. Specifically, the comprehensive experimental results demonstrate that  \textbf{TransRef}  learns to well establish the relationship between the input image and reference information rather than solely relying on the prior knowledge of large-scale training data, despite the huge domain gaps between the training dataset and the application data. 

\section{Acknowledgement}
\label{sec:acknowledgement}
This work was supported by National Natural Science Foundation of China (62202349, 62371351), and Young Elite Scientists Sponsorship Program by CAST (2023QNRC001).

\bibliographystyle{elsarticle-num} 
\bibliography{cas-refs}





% \end{thebibliography}

\end{document}